\documentclass[11pt]{article}

 \usepackage{natbib}
 \usepackage{amssymb}
 \usepackage{amsbsy}
 \usepackage{amscd}
 \usepackage{amsmath}
 \usepackage{amsthm}

\newtheorem{Theorem}{{\bf Theorem}}
\newtheorem{Lemma}{{\bf Lemma}}

\newtheorem*{keywords}{\bf Keywords}
\newcommand{\RR}{\mathbb R}

\newcommand{\EE}{\mathbb E}

\title{Bayesian Cross Validation and WAIC for Predictive Prior Design in Regular Asymptotic Theory}

\author{
Sumio Watanabe \\
Department of Computational Intelligence and Systems Science\\
Tokyo Institute of Technology\\
Mailbox G5-19, 4259 Nagatsuta, Midori-ku\\
Yokohama, Japan 226-8502 \\
swatanab@dis.titech.ac.jp \\
}

\begin{document}

\maketitle

\begin{abstract}
Prior design is one of the most important problems in both statistics and machine learning. 
The cross validation (CV) and the widely applicable information criterion (WAIC) 
are predictive measures of the Bayesian estimation, however, it has been difficult to 
apply them to find the optimal prior because their mathematical properties in prior evaluation have been unknown and 
the region of the hyperparameters is too wide to be examined.
In this paper, we derive a new formula by which the theoretical relation among CV, WAIC, and the generalization loss is 
clarified and the optimal hyperparameter can  be directly found. 

By the formula, three facts are clarified about predictive prior design. 
Firstly, CV and WAIC have the same second order asymptotic expansion, hence they are 
asymptotically equivalent to each other as the optimizer of the hyperparameter. Secondly, 
the hyperparameter which minimizes CV or WAIC makes
the average generalization loss to be minimized asymptotically but does not the random generalization loss. 
And lastly, by using the mathematical relation between priors, the variances of the optimized hyperparameters by CV and WAIC are made 
smaller with small computational costs. 
Also we show that the optimized hyperparameter by DIC or the marginal likelihood does not minimize the average or random 
generalization loss in general. 
\end{abstract}

\begin{keywords}
Hyperparameter, Cross validation, WAIC, DIC,  marginal likelihood
\end{keywords}

\section{Introduction}

In statistics and machine learning, a method how to design 
a prior distribution is one of the most important problems. 
It is well known that Bayesian estimation or some regularization techniques are useful in practical 
statistical problems, however, its performance strongly depends on the prior, 
hence we need the theoretical foundation which enables us to evaluate the chosen prior. 

Sometimes the parameter in the prior is called a hyperparameter, and 
the prior design problem results in the method how to choose the optimal hyperparameter. 
Historically, it was proposed that a hyperparameter is optimized by maximization of the
marginal likelihood \citep{Good,ABIC}. This method is one of the rational procedures because
it can be understood as the maximum likelihood method for the marginal distribution, however, 
the optimal prior for this criterion does not minimize the average generalization loss in general. 
In this paper, we study predictive prior design, in other words, a method 
how to choose a prior so as to minimize the average generalization loss. 

The generalization loss can be estimated by the cross validation (CV) and 
information criteria. In Bayesian estimation, the leave-one-out cross validation
can be approximated by using the important sampling cross validation \citep{Gelfand,Vehtari}, 
whose statistical property was studied in Bayesian statistics \citep{Peruggia,Epifani,Vehtari2}. 
The deviance information criterion (DIC)
was proposed  for the case when the true distribution is realizable by a 
statistical model and the posterior is a normal distribution \citep{DIC,DIC2}. 
For general cases when  the true distribution 
may be unrealizable or the posterior may not be a normal distribution, the widely 
applicable information criterion (WAIC) was proposed based on singular learning theory \citep{NC2001,Cambridge2009,NN2010} and it was proved that
WAIC is asymptotically equivalent to the leave-one-out cross validation \citep{JMLR2010}.
Both CV and WAIC are studied by using 
the Hamiltonian Monte Carlo methods and its improved algorithm using No-U-Turn dynamics \citep{Gelman,Gelman2,Vehtari2}. 

From the predictive point of view, both CV and information criteria have three problems. 
The first is a theoretical problem about consistency. 
Both CV and information criteria are asymptotically unbiased estimators of the average generalization loss,
in other words, 
their expectation values are asymptotically equal to that of the generalization loss. However, in general, 
the minimum point of a random function is not equal to that of the average function. 
Therefore, it has been unknown whether minimization of CV or information criteria is asymptotically equal to 
minimization of the average generalization loss or not. 
In this paper, we prove that, if a statistical model is regular, minimization of CV or WAIC makes
the average generalization loss to be minimized asymptotically, whereas minimization of DIC or maximization of
the marginal likelihood does not even asymptotically. 
 
The second is a problem about the difference between the random and average generalization losses. 
The former is the random variable which depends on a given set of training samples,
whereas the latter is the expectation value over all training sets taken from the true distribution.
We show in this paper that the hyperparameter that minimizes the random generalization loss does not converge to that
of the average generalization loss. It follows that, 
although the optimal hyperparameter for the minimum CV or WAIC asymptotically minimizes the
average generalization loss, it does not the random generalization loss. 

The last is a practical problem. 
When CV or information criteria is employed, it is not easy to determine the region of candidate hyperparameters,
because we do not know whether the optimal hyperparameter exists, or, if it does, where it is. 
Moreover, after the region is determined, in order to compare CV or information criteria 
the posterior distributions are required for all candidate hyperparameters, resulting in heavy computational costs. 
The new formula obtained in this paper enables us to directly estimate CV and WAIC as a function of a hyperparameter, 
hence the optimal hyperparameter can be found without comparing candidate hyperparameters. 
Also we show a method by which the variances of the chosen hyperprameters by CV and WAIC are made smaller
using the new formula. 

This paper consists of eight sections. In the second and third sections, we define the basic definitions
in Bayesian statistical learning and introduce the main results of this paper. 
In the fourth section, we study two examples. 
The fifth and sixth chapters are devoted to the proofs of  the basic lemmas and  
the main theorems respectively.
In the seventh section, several points of the main results are discussed, and in the last section, 
we conclude the paper with the problem for the future study. 

\section{Basic Definitions in Bayesian Statistical Learning} 

In this section, we introduce the basic definitions in Bayesian statistical learning. 
Let $q(x)$ be a probability density function on  the $N$ dimensional real
Euclidean space $\RR^{N}$, and $X_1,X_2,...,X_n$ be random variables 
on $\RR^{N}$ which are independently subject to $q(x)$. 
The probability density function $q(x)$ is sometimes referred to as a 
true distribution. A training set is denoted by
$
X^n=(X_1,X_2,...,X_n)
$, where $n$ is the number of training samples. 
The average $\EE[\;\;]$ shows the expectation value overall training sets $X^n$.
A statistical model or a learning machine is defined by 
$p(x|w)$ which is a probability density function of $x\in \RR^N$ for a given 
parameter $w\in W\subset \RR^{d}$. A nonnegative function $\varphi(w)$ on the
parameter set $W$ is
called a  prior distribution. If it satisfies
\begin{equation}
\int \varphi(w)dw =1,\label{eq:prior_one}
\end{equation}
then $\varphi(w)$ is called to be proper.
In this paper, we study both proper and improper priors, hence eq.(\ref{eq:prior_one}) 
does not hold in general. 
The posterior distribution is defined by
\begin{equation}
p(w|X^n)=\frac{1}{Z(\varphi)}\varphi(w)\prod_{i=1}^n p(X_i|w),
\end{equation}
where $Z(\varphi)$ is a normalizing constant. 
\[
Z(\varphi)=\int \varphi(w)\prod_{i=1}^n p(X_i|w) dw. 
\]
We assume that $Z(\varphi)$ is finite with probability one. If $\varphi(w)$ is proper, then
$Z(\varphi)$ is equal to the marginal likelihood. 
The expectation value of a given function $f(w)$ over the posterior distribution 
is denoted by
\begin{equation}
\EE_{\varphi}[f(w)]=\int f(w) p(w|X^n)dw.
\end{equation}
The predictive distribution is the average of a statistical model over the posterior distribution, 
\begin{equation}
p(x|X^n)=\EE_{\varphi}[p(x|w)].
\end{equation}
The random generalization loss is defined by 
\begin{equation}
G(\varphi)=-\int q(x)\log p(x|X^n)dx.\label{eq:G(varphi)}
\end{equation}
Note that the random variable $G(\varphi)$ depends on the training set $X^n$.
The average generalization loss is defined by $\EE[G(\varphi)]$.
 In this paper, we show that  the random variable $G(\varphi)$ has a different behavior 
from its expectation value $\EE[G(\varphi)]$ as a functional of $\varphi(w)$ even asymptotically. 
The  Bayesian leave-one-out cross validation (CV)  is defined by 
\begin{eqnarray}
{\rm CV}(\varphi)&=& -\frac{1}{n}\sum_{i=1}^{n}\log p(X_i|X^n\setminus X_i) \label{eq:CVoriginal} \\
&=& \frac{1}{n}\sum_{i=1}^n\log \EE_{\varphi}\Bigl[\frac{1}{p(X_i|w)}\Bigr],\label{eq:ISCV}
\end{eqnarray}
where $X^n\setminus X_i$ is a set of training samples leaving $X_i$ out.
A  calculation method of CV by eq.(\ref{eq:ISCV}) using the posterior distribution by 
the Markov chain Monte Carlo method is sometimes called the important sampling cross validation. 
The training error $T(\varphi)$ and the functional variance $V(\varphi)$ are 
respectively defined by
\begin{eqnarray}
T(\varphi)&=&-\frac{1}{n}\sum_{i=1}^n\log p(X_i|X^n)=-\frac{1}{n}\sum_{i=1}^n\log\EE_{\varphi}[p(X_i|w)],\\
V(\varphi)&=&\sum_{i=1}^{n}\{\EE_{\varphi}[(\log p(X_i|w))^2]-\EE_{\varphi}[\log p(X_i|w)]^2\}.
\end{eqnarray}
Then the widely applicable information criterion (WAIC) is defined by
\begin{equation}
{\rm WAIC}(\varphi)=T(\varphi)+\frac{1}{n}V(\varphi).
\end{equation}
For a real number $\alpha$, the functional cumulant function is defined by
\begin{equation}\label{eq:functional_cumulant}
F_{cum}(\alpha)=\frac{1}{n}\sum_{i=1}^n\log\EE_{\varphi}[p(X_i|w)^{\alpha}].
\end{equation}
Then, as is shown in \citep{JMLR2010},
\begin{eqnarray}
{\rm CV}(\varphi)&=&F(-1),\label{eq:CV_F}\\
T(\varphi)&=&-F(1),\label{eq:T_F}\\
V(\varphi)&=&n F''(0),\label{eq:V_F}\\
{\rm WAIC}(\varphi)&=&-F(1)+F''(0).\label{eq:WAIC_F}
\end{eqnarray}
In the previous papers, we proved by singular learning theory that, even if a true distribution is unrealizable by a statistical model or even if
the posterior distribution is not the normal distribution, 
\begin{eqnarray*}
\EE[{\rm CV}(\varphi)]&=&\EE[G(\varphi)]+O(\frac{1}{n^2}),\\
\EE[{\rm WAIC}(\varphi)]&=&\EE[G(\varphi)]+O(\frac{1}{n^2}),\\
{\rm WAIC}(\varphi)&=&{\rm CV}(\varphi)+O_p(\frac{1}{n^2}).
\end{eqnarray*}
However, it has been left unknown whether minimization of ${\rm CV}(\varphi)$ or 
${\rm WAIC}(\varphi)$ with respect to a prior $\varphi$ 
is asymptotically equivalent to minimization of $G(\varphi)$ and $\EE[G(\varphi)]$ or not. 
In this paper, 
we prove in Theorem 1 that, if a statistical model satisfies the several regularity conditions, 
minimization of ${\rm CV}(\varphi)$ or ${\rm WAIC}(\varphi) $
is asymptotically equivalent to minimization of $\EE[G(\varphi)]$ but not to $G(\varphi)$. 

In  the hyperparameter optimization problem, two alternative methods are well known. 
The former is maximization of the marginal likelihood or equivalently minimization of 
the free energy or the minus log marginal likelihood,
\begin{equation}\label{eq:FF}
F_{free}(\varphi)= -\log \int \varphi(w)\prod_{i=1}^n p(X_i|w) dw+\log\int\varphi(w)dw.
\end{equation}
In order to use this method, the integration of $\varphi(w)$ should be finite, because, if it is not finite, $F_{free}(\varphi)$ can
not be defined.  
The latter is the deviance informaiton criterion (DIC), 
\begin{equation}\label{eq:DIC}
{\rm DIC}(\varphi)=\frac{1}{n}\sum_{i=1}^{n}\{-2\EE_{\varphi}[\log p(X_i|w)]+\log p(X_i|\EE_{\varphi}[w])\}.
\end{equation}
In this method a prior may be improper like as CV and WAIC. In this paper, we show that the hyperparameter 
which minimizes $F_{free}(\varphi)$ or ${\rm DIC}(\varphi)$ does not minimize either $\EE[G(\varphi)]$ or $G(\varphi)$
even asymptotically in general.

\section{Main Results}

\subsection{Definitions and Conditions}

In this section, we introduce several notations, regularity conditions, and definitions of 
mathematical relations between priors.

The set of parameters $W$ is assumed to be an open subset of $\RR^d$.
In this paper, $\varphi_0(w)$ is an arbitrary fixed prior and $\varphi(w)$ is a candidate prior which will be optimized. 
We assume that, for an arbitrary $w\in W$, $\varphi_0(w)>0$  and $\varphi(w)>0$.
We do not assume that they are proper. 
The main purpose of this paper is make a new formula which enables us to directly estimate 
${\rm CV}(\varphi)-{\rm CV}(\varphi_0)$ and ${\rm WAIC}(\varphi)-{\rm WAIC}(\varphi_0)$.

The prior ratio function $\phi(w)$ is denoted by 
\[
\phi(w)=\frac{\varphi(w)}{\varphi_0(w)}.
\]
If $\varphi_0(w)\equiv 1$, then $\phi(w)=\varphi(w)$. 
The empirical log loss function and the maximum {\it a posteriori} (MAP) estimator $\hat{w}$ 
are respectively defined by 
\begin{eqnarray}\label{eq:L(w)}
L(w) &=&-\frac{1}{n}\sum_{i=1}^n\log p(X_i|w)-\frac{1}{n}\log \varphi_0(w), \\
\hat{w}&=& \mbox{arg}\min_{w\in W} L(w),\label{eq:MAP}
\end{eqnarray}
where either $L(w)$ or $\hat{w}$ does not depend on $\varphi(w)$. 
If $\varphi_0(w)\equiv 1$, then $\hat{w}$ is equal to the maximum likelihood estimator (MLE). 
The average log loss function and the parameter that minimizes it are respectively defined by 
\begin{eqnarray}
{\cal L}(w)&=&-\int q(x)\log p(x|w)dx,\\
w_0&=&\mbox{arg}\min_{w\in W} {\cal L}(w).
\end{eqnarray}
In this paper we use the following notations for simple description. 
\vskip5mm
\noindent{\bf Notations.}\\
(1)  A parameter is denoted by $w=(w^1,w^2,...,w^k,...,w^d)\in\RR^{d}$. Remark that $w^{k}$ means
the $k$th element of $w$, which does not mean $w$ to the power of $k$. \\
(2) For a given real function $f(w)$ and nonnegative integers $k_1,k_2,...,k_m$, we define 
\begin{equation}
f_{k_1k_2\cdots k_m}=f_{k_1k_2\cdots k_m}(w)=
\frac{\partial^m f}
{\partial w^{k_1}\partial w^{k_2}\cdots \partial w^{k_m}}(w). \label{eq:derivative}
\end{equation}
(3) We adopt Einstein's summation convention and $k_1,k_2,k_3,...$ are
used for such suffices. For example,
\[
X_{k_1k_2}Y^{k_2k_3}=\sum_{k_2=1}^{d}X_{k_1k_2}Y^{k_2k_3}. 
\]
In other words, if a suffix $k_i$ appears both upper and lower, it means automatic summation
over $k_i=1,2,...,d$. In this paper, for each $k_1,k_2$,  $X^{k_1k_2}=X^{k_1}_{k_2}=X_{k_1k_2}$. 
\vskip5mm
In order to prove the main theorem, we need the regularity conditions. In this paper, we do not
study singular learning machines. 
\vskip5mm
\noindent{\bf Regularity Conditions.}\\
(1) ({\bf Parameter Set})  The parameter set $W$ is an open set in $R^{d}$. \\
(2) ({\bf Smoothness of Models}) The functions $\log \varphi(w)$, $\log \varphi_0(w)$, and $\log p(x|w)$ are 
$C^{\infty}$-class functions of $w\in W$, in other words, they are infinitely many times differentiable. \\
(3) ({\bf Identifiability of Parameter}) There exists a unique $w_0\in W$ which minimizes the average log loss function ${\cal L}(w)$. 
There exists a unique $\hat{w}\in  W$ which minimizes 
$L(w)$ with probability one. It is assumed that the convergence in probability $\hat{w}\rightarrow w_0$ $(n\rightarrow\infty)$ holds. \\
(4) ({\bf Regularity Condition}) The matrix ${\cal L}_{k_1k_2}(w_0)$ is invertible. Also
the matrix $L_{k_1k_2}(w)$ is invertible 
for almost all $w$ in a neighborhood of $w_0$ 
with probability one. Let $J^{k_1k_2}(w)$ be the inverse matrix of $L_{k_1k_2}(w)$.  \\
(5) ({\bf Well-Definedness and Concentration of Posterior}) We assume that, for an arbitrary  $|\alpha|\leq 1$ and $j=1,2,...,n+1$,
\begin{eqnarray}
\EE_{X_{n+1}}\EE\Bigl[\Bigl{|}\log \EE_{\varphi}[p(X_j|w)^{\alpha}]\Bigr{|}\Bigr]<\infty.\label{eq:finite}
\end{eqnarray}
The same inequality as eq.(\ref{eq:finite}) holds for $\varphi_0(w)$ instead of $\varphi(w)$. 
Let $Q(X^n,w)$ be an arbitrary finite times product of
\begin{eqnarray*}
&& (\log\varphi(w))_{k_1k_2\cdots k_p},\\
&&(\log \varphi_0(w))_{k_1k_2\cdots k_q},\\
&&\frac{1}{n}\sum_{i=1}^n \prod (\log p(X_i|w))_{k_1k_2\cdots k_r}, \\
&&(J^{k_1k_2}(w))_{k_1k_2\cdots k_s}, \\
&& w^{k_1},  
\end{eqnarray*}
where $|\alpha|\leq 1$, $p,q,r,s\geq 0$ and $\prod$ shows a finite product of a combination $(k_1,k_2,..,k_r)$. 
Let 
\begin{equation}\label{eq:Wepsilon}
W(\varepsilon)=\{w\in W ;|w-\hat{w}|<n^{\varepsilon-1/2}\}.
\end{equation}
It is assumed that there exists $\varepsilon>0$, for an arbitrary such product $Q(X^n,w)$, 
\begin{eqnarray}
\EE[\sup_{W(\varepsilon)} |Q(X^n,w)|]<\infty,\label{eq:condition1}\\
\EE[ Q(X^n,\hat{w})] \rightarrow \EE[ Q(X^n,w_0) ],\label{eq:condition2}
\end{eqnarray}
and that, for arbitrary $|\alpha|\leq 1$ and $\beta>0$,
\begin{eqnarray}
&& \frac{
\EE_{\varphi}[Q(X^n,w)p(X_j|w)^{\alpha}]
}{
\EE_{\varphi}[p(X_j|w)^{\alpha}]
} \nonumber \\
&& =\Bigl (1+o_p\bigl(\frac{1}{n^{\beta}}\bigr)\Bigr)
\frac{
\displaystyle \int_{W(\varepsilon)} Q(X^n,w)
p(X_j|w)^{\alpha}\prod_{i=1}^{n}p(X_i|w)\varphi(w)dw
}{
\displaystyle \int_{W(\varepsilon)} 
p(X_j|w)^{\alpha} \prod_{i=1}^{n}p(X_i|w)\varphi(w)dw
}\label{eq:condition3},
\end{eqnarray}
where $o_p(1/n^{\beta})$ satisfies $n^{\beta}\EE[|o_p(1/n^{\beta})|]\rightarrow 0$. 
Also we assume  that the same equation as eq.(\ref{eq:condition3})  holds for $\varphi_0(w)$ instead of $\varphi(w)$. 
\vskip5mm\noindent
{\bf Explanation of Regularity Condition}. (1) In this paper, we assume that $p(x|w)$ is regular at $w_0$, that is to say, 
the second order matix ${\cal L}_{k_1k_2}(w_0)$ is positive definite. If this condition is not satisfied, then such $(q(x),p(x|w))$ is called 
singular. The results of this paper do not hold for singular learning machines. \\
(2) Conditions eq.(\ref{eq:condition1}) and eq.(\ref{eq:condition2}) ensure 
the finiteness of the expectation values and concentration of the posterior distribution. 
The condition of the concentration, eq.(\ref{eq:condition3}), is set by the following mathematical reason. 
Let $S(w)$ be a function which takes the minimum value $S(\hat{w})=0$ at $w=\hat{w}$. If $S_{k_1k_2}(\hat{w})$ is
positive definite, then by using the saddle point approximation in the neighborhood of $\hat{w}$, 
\[
\exp(-nS(w))\approx \exp\Bigl(-\frac{n}{2}S_{k_1k_2}(\hat{w})(w-\hat{w})^{k_1}(w-\hat{w})^{k_2}\Bigr),
\]
hence the orders of integrations inside and outside of $W(\epsilon)$ are respectively given by 
\begin{eqnarray}
\int_{W(\epsilon)} \exp(-n S(w))dw&=&O(1/n^{d/2}),\nonumber \\
\int_{W\setminus W(\epsilon)}\exp(-nS(w))dw &=&O(\exp(-n^{\varepsilon})).\nonumber 
\end{eqnarray}
Therefore the integration over  $W\setminus W(\varepsilon)$ conveges to zero faster than that over $W(\epsilon)$
as $n\rightarrow\infty$.

\vskip3mm
\noindent{\bf Definition.} (Empirical Mathematical Relations between Priors) 
The empirical mathematical relation between two priors $\varphi(w)$ and $\varphi_0(w)$ at a parameter 
$w$  is defined by 
\begin{eqnarray}
M(\phi,w)&=&
A^{k_1k_2}
(\log \phi)_{k_1}(\log\phi)_{k_2} + B^{k_1k_2}
(\log \phi)_{k_1k_2}  +
C^{k_1}(\log \phi)_{k_1},\label{eq:MMM}
\end{eqnarray}
where $\phi(w)=\varphi(w)/\varphi_0(w)$ and 
\begin{eqnarray*}
J^{k_1k_2}(w)&=&\mbox{Inverse matrix of } L_{k_1k_2}(w),\\
A^{k_1k_2}(w)&=&\frac{1}{2}J^{k_1k_2}(w), \\
B^{k_1k_2}(w)&=&\frac{1}{2}(J^{k_1k_2}(w)+J^{k_1k_3}(w)J^{k_2k_4}(w)F_{k_3,k_4}(w)),\\
C^{k_1}(w)&=& J^{k_1k_2}(w)J^{k_3k_4}(w)F_{k_2k_4,k_3} (w)
-\frac{1}{2}J^{k_1k_2}(w)J^{k_3k_4}(w)L_{k_2k_3k_4}(w),\\
&& -\frac{1}{2}J^{k_1k_2}(w)J^{k_3k_4}(w)J^{k_5k_6}(w)L_{k_2k_3k_5}(w)F_{k_4,k_6}(w),
\end{eqnarray*}
where $L_{k_1k_2}(w)$ and $L_{k_1k_2k_3}(w)$ are the second and third derivatives of $L(w)$ respectively 
as defined by eq.(\ref{eq:derivative}) and 
\begin{eqnarray*}
F_{k_1,k_2}(w)&=&\frac{1}{n}\sum_{i=1}^n (\log p(X_i|w))_{k_1}( \log p(X_i|w))_{k_2},\\
F_{k_1k_2,k_3}(w)&=&\frac{1}{n}\sum_{i=1}^n (\log p(X_i|w))_{k_1k_2}
( \log p(X_i|w))_{k_3}.
\end{eqnarray*}
\vskip3mm\noindent
{\bf Remark.} Note that neither $A^{k_1k_2}(w)$, $B^{k_1k_2}(w)$, nor $C^{k_1}(w)$ 
depends on a candidate prior $\varphi(w)$. 
\vskip3mm
\noindent{\bf Definition.} (Average Mathematical Relations between Priors) 
The average mathematical relation ${\cal M}(\phi,w)$ 
is defined by the same manner as eq.(\ref{eq:MMM}) by using
\begin{eqnarray}
{\cal J}^{k_1k_2}(w)&=&\mbox{Inverse matrix of } {\cal L}_{k_1k_2}(w),\\
{\cal L}_{k_1k_2}(w)&=&\int (-\log p(x|w))_{k_1k_2}q(x)dx,\\
{\cal L}_{k_1k_2k_3}(w)&=&\int (-\log p(x|w))_{k_1k_2k_3}q(x)dx,\\
{\cal F}_{k_1,k_2}(w)&=& \int  (\log p(x|w))_{k_1}
( \log p(x|w))_{k_2}q(x)dx,\\
{\cal F}_{k_1k_2,k_3}(w)&=&  \int  (\log p(x|w))_{k_1k_2}
( \log p(x|w))_{k_3}q(x)dx,
\end{eqnarray}
instead of $J^{k_1k_2}(w)$, 
$L_{k_1k_2}(w)$, $L_{k_1k_2k_3}(w)$, 
$F_{k_1,k_3}(w)$, and 
$F_{k_1k_2,k_3}(w)$ respectively. \\
The self-average mathematical relation $\langle M\rangle (\phi,w)$ 
is defined by the same manner as $M(\phi,w)$ by using 
\begin{eqnarray}
\langle J^{k_1k_2}\rangle (w)&=&\mbox{Inverse matrix of } \langle L_{k_1k_2}\rangle (w), \\
\langle L_{k_1k_2}\rangle(w) &=&\int (-\log p(x|w))_{k_1k_2}p(x|w)dx,\\
\langle L_{k_1k_2k_3}\rangle(w) &=&\int (-\log p(x|w))_{k_1k_2k_3}p(x|w)dx,\\
\langle F_{k_1,k_2}\rangle(w)  &=& \int  (\log p(x|w))_{k_1}
( \log p(x|w))_{k_2}p(x|w)dx,\\
\langle F_{k_1k_2,k_3}\rangle(w)  &=&  \int  (\log p(x|w))_{k_1k_2}
( \log p(x|w))_{k_3}p(x|w)dx,
\end{eqnarray}
instead of $J^{k_1k_2}(w) $, 
$L_{k_1k_2}(w) $, $L_{k_1k_2k_3}(w) $, 
$F_{k_1,k_3}(w) $, and 
$F_{k_1k_2,k_3}(w) $ respectively.
\vskip3mm\noindent
{\bf Remark.} In the self-average case, it holds that $\langle L_{k_1k_2}\rangle(w) = \langle F_{k_1,k_2}\rangle(w)$, hence
 $\langle M \rangle (\phi,w)$ 
can be calculated by the same manner as eq.(\ref{eq:MMM}) by using 
\begin{eqnarray*}
\langle A^{k_1k_2}\rangle(w) &=&\frac{1}{2}\langle J^{k_1k_2}\rangle(w),  \\
\langle B^{k_1k_2}\rangle(w) &=&\langle J^{k_1k_2}\rangle(w), \\
\langle C^{k_1}\rangle(w) &=&\langle J^{k_1k_2}\rangle(w) \langle J^{k_3k_4}\rangle(w)  \langle F_{k_2k_4,k_3} \rangle(w) 
-\langle J^{k_1k_2}\rangle(w)  \langle J^{k_3k_4}\rangle (w) \langle L_{k_2k_3k_4}\rangle(w) .
\end{eqnarray*}
instead of $A^{k_1k_2}(w) $, $B^{k_1k_2}(w) $ and $C^{k_1}(w) $.

\subsection{Main Theorem}

The following  is the main result of this paper. 
\begin{Theorem} Assume the regularity conditions (1), (2), ..., and (5).  
Let $M(\phi,w)$ and ${\cal M}(\phi,w)$ be the empirical and average mathematical relations between $\varphi(w)$ 
and $\varphi_0(w)$. Then 
\begin{eqnarray}
{\rm CV}(\varphi)&=&{\rm CV}(\varphi_0)+\frac{M(\phi,\hat{w})}{n^2}+O_p(\frac{1}{n^3}),\label{eq:CV}\\
\EE[{\rm CV}(\varphi)]&=&\EE[{\rm CV}(\varphi_0)]+\frac{{\cal M}(\phi,w_0)}{n^2}+O(\frac{1}{n^3}),\label{eq:ECV}\\
{\rm WAIC}(\varphi)&=&{\rm WAIC}(\varphi_0)+\frac{M(\phi,\hat{w})}{n^2}+O_p(\frac{1}{n^3}), \label{eq:WAIC}\\
\EE[{\rm WAIC}(\varphi)]&=&\EE[{\rm WAIC}(\varphi_0)]+\frac{{\cal M}(\phi,w_0)}{n^2}+O(\frac{1}{n^3}),
\label{eq:EWAIC} \\
{\rm CV}(\varphi)&=&{\rm WAIC}(\varphi)+O_p(\frac{1}{n^3}),\label{eq:CV_WAIC}
\end{eqnarray}
and 
\begin{eqnarray}
M(\phi,\hat{w})&=&{\cal M}(\phi,w_0)+O_p(\frac{1}{n^{1/2}}), \label{eq:MM}\\
M(\phi,\EE_{w}[w])&=&M(\phi,\hat{w})+O_p(\frac{1}{n}), \label{eq:MM2}\\
\EE[M(\phi,\hat{w})]&=&{\cal M}(\phi,w_0)+O(\frac{1}{n}).\label{eq:EMM}
\end{eqnarray}
On the other hand, 
\begin{eqnarray}
G(\varphi)&=&G(\varphi_0)+\frac{1}{n}(\hat{w}^{k_1}-(w_0)^{k_1})(\log\phi)_{k_1}(\hat{w})+O_p(\frac{1}{n^2})\nonumber\\
&=& G(\varphi_0)+O_p(\frac{1}{n^{3/2}}),\label{eq:G} \\
\EE[G(\varphi)]&=&\EE[G(\varphi_0)]+\frac{{\cal M}(\phi,w_0)}{n^2}+O(\frac{1}{n^3}). \label{eq:EG}
\end{eqnarray}
\end{Theorem}
\vskip3mm\noindent
From Theorem 1, the five mathematical facts are derived. \\
(1) Assume that a prior $\varphi(w)$ has a hyperparameter. 
Let $h(f)$ be the hyperparameter that minimizes $f(\varphi)$. 
By eq.(\ref{eq:CV}) and eq.(\ref{eq:WAIC}), 
$h({\rm CV})$ and $h({\rm WAIC})$ can be directly found by minimizing
the empirical methematical relation $M(\phi,\hat{w})$ asymptotically. 
By eq.(\ref{eq:MM})  and eq.(\ref{eq:EG}), $h({\rm CV})$ and $h({\rm WAIC})$ is asymptotically equal to
$h(\EE[G])$. 
Remark that ${\rm CV}(\varphi) $, ${\rm WAIC}(\varphi)$, and $G(\varphi)$ may be unbounded 
or may not have a minimum value as a 
function of a hyperparameter if the set of all hyperparameters is not compact. 
By using $M(\phi,\hat{w})$, we can examine whether they have a minimum value or not. 
The divergence phenomenon of ${\rm CV}(\varphi)$ and ${\rm WAIC}(\varphi)$ as functions on a noncompact set of hyperparameters 
 is discussed in Section \ref{section:div}. \\
(2)  The variance of $h({\rm CV})$ is 
asymptotically equal to that of $h({\rm WAIC})$, however, they may be different when the number of
training samples are finite. \\
(3) In calculation of the mathematical relation $M(\phi,\hat{w})$, the MAP estimator $\hat{w}$ can be replaced by
the posterior average parameter $\EE_{w}[w]$ asymptotically.\\
(4) By eq.(\ref{eq:G}), the variance of the random generalization loss $G(\varphi)-G(\varphi_0)$ 
is  larger than those of ${\rm CV}(\varphi)-{\rm CV}(\varphi_0)$ and ${\rm WAIC}(\varphi)-{\rm WAIC}(\varphi_0)$. 
Neither $h({\rm CV})$ nor $h({\rm WAIC})$ 
minimizes the random generalization loss $G(\varphi)$ in general. \\
(5) It was proved in (Watanabe, 2010) that $\EE[G(\varphi_0)]=d/(2n)+o(1/n)$, where $d$ is the dimension of
the parameter set. 
Assume that there exist finite sets of real values $\{d_{k}\}$ and $\{\gamma_k\}$, where $\gamma_k>1$, such that
\[
\EE[G(\varphi_0)]=\frac{d}{2n}+\sum_{k}\frac{d_k}{n^{\gamma_k}}+o(\frac{1}{n^2}).
\]
Since $\EE[{\rm CV}(\varphi_0)]$ of $X^n$ is equal to $\EE[G(\varphi_0)]$ of $X^{n-1}$ and 
\[
\frac{1}{n-1}-\frac{1}{n}=\frac{1}{n^2}+o(\frac{1}{n^2}),
\]
it immediately follows from Theorem 1 that
\begin{eqnarray}
\EE[G(\varphi)]&=&\EE[G(\varphi_0)]+\frac{{\cal M}(\phi,w_0)}{n^2}+o(\frac{1}{n^2}),\\
\EE[{\rm CV}(\varphi)]&=&\EE[G(\varphi_0)]+\frac{d/2+{\cal M}(\phi,w_0)}{n^2}+o(\frac{1}{n^2}),\\
\EE[{\rm WAIC}(\varphi)]&=&\EE[G(\varphi_0)]+\frac{d/2+{\cal M}(\phi,w_0)}{n^2}+o(\frac{1}{n^2}).
\end{eqnarray}

\begin{Theorem} Assume the regularity conditions (1), (2), ..., and (5).  
If there exists a parameter $w_0$ such that $q(x)=p(x|w_0)$, then  
\begin{eqnarray}
\langle M\rangle(\phi,\hat{w})&=& M(\phi,\hat{w})+O_p(\frac{1}{\sqrt{n}}),\\
\langle M\rangle(\phi,\hat{w})&=& {\cal M}(\phi,w_0)+O_p(\frac{1}{\sqrt{n}}).
\end{eqnarray}
\end{Theorem}
\vskip3mm\noindent
By Theorem 2, if the true distribution is realizable by a  statistical model or a learning machine, then the empirical mathematical 
relation can be replaced by its self-average. The variance of the self-average mathematical relation is
often smaller the original one, hence the variance of the estimated hyperparameter by using the self-average is made smaller.

Based on Theorem 1 and 2, we define new information criteria for hyperparameter optimization,
the widely applicable information criterion for a regular case and a regular case using self-average, 
\begin{eqnarray}
{\rm WAICR}&=&\frac{M(\phi,\hat{w})}{n^2},\\
{\rm WAICRS}&=&\frac{\langle M\rangle(\phi,\hat{w})}{n^2},
\end{eqnarray}
where $\hat{w}$ can be replaced by $\EE_{\varphi_0}[w]$. The optimal hyperparameter for 
predictive prior design 
can be directly found by minimization of these criteria if they have the minimum points. 
\section{Examples}

\subsection{Normal Distribution}

A simple but nontrivial example is a normal distribution whose mean and standard deviation 
are $(m,1/s)$, 
\begin{eqnarray}
p(x|m,s)&=&\sqrt{\frac{s}{2\pi}}\exp\Bigl(-\frac{s}{2}(x-m)^2\Bigr).\label{eq:ND}
\end{eqnarray}
For a prior distribution, we study
\begin{eqnarray}\label{eq:Prior}
\varphi(m,s|\lambda,\mu,\epsilon)&=&\exp\Bigl(-\frac{\lambda sm^2+\varepsilon s}{2}\Bigr)s^{\mu},
\end{eqnarray}
where $(\lambda,\mu,\varepsilon)$ is a set of hyperparameters.
Note that the prior is improper in general. 
If $\lambda>0$, $\mu>-1/2$ and $\varepsilon>0$, the prior can be made proper 
by 
\[
\Phi(m,s|\lambda,\mu,\varepsilon)=\frac{1}{C}\varphi(m,s|\lambda,\mu,\varepsilon),
\]
where
\[
C=\sqrt{\frac{2\pi}{\lambda}}(\varepsilon/2)^{-\mu-1/2}\Gamma(\mu+1/2).
\]
We use a fixed prior as $\varphi_0(m,s)\equiv 1$, then 
the empirical log loss function is given by
\begin{eqnarray}
L(m,s)&=& -\frac{1}{2}\log \frac{s}{2\pi}+\frac{s}{2n}\sum_{i=1}^{n}(X_i-m)^2.
\end{eqnarray}
Let $M_j= (1/n)\sum_{i=1}^n(X_i-\hat{m})^j$ $(j=2,3,4)$. 
The MAP estimator is equal to the MLE $ \hat{w}=(\hat{m},\hat{s})$, where $\hat{s}^2=1/M_2$, resulting that  
\begin{eqnarray}
A^{k_1k_2}(\hat{w})&=&\left(
\begin{array}{cc}
1/(2\hat{s})&0 \\
0&\hat{s}^2
\end{array}\right),\\
B^{k_1k_2}(\hat{w})&=&
\left(
\begin{array}{cc}
1/\hat{s}&-\hat{s}^2M_3/2 \\
-\hat{s}^2M_3/2&(\hat{s}^2+ \hat{s}^4M_4)/2
\end{array}\right),
\\
C^{k_1}(\hat{w})&=&
(0, \hat{s}+\hat{s}^3M_3).
\end{eqnarray}
Also the self-average mathematical relation is given by 
\begin{eqnarray}
\langle A^{k_1k_2}\rangle (\hat{w})&=&\left(
\begin{array}{cc}
1/(2\hat{s})&0 \\
0&\hat{s}^2
\end{array}\right),\\
\langle B^{k_1k_2}\rangle(\hat{w})&=&
2 \langle A^{k_1k_2}\rangle (\hat{w}),
\\
\langle C^{k_1}\rangle(\hat{w})&=&(0, \hat{s}).
\end{eqnarray}
The prior ratio function is $\phi(w)=\varphi(w)$, hence
the derivatives of the log prior ratio are 
\begin{eqnarray*}
(\log\phi)_1(\hat{w})&=&-\lambda  \hat{s}\hat{m},\\
(\log\phi)_2(\hat{w})&=&-\frac{\lambda \hat{m}^2}{2}+\mu/ \hat{s} -\frac{\varepsilon}{2},\\
(\log\phi)_{11}(\hat{w})&=&-\lambda  \hat{s},\\
(\log\phi)_{12}(\hat{w})&=&-\lambda \hat{m},\\
(\log\phi)_{22}(\hat{w})&=&-\mu/ \hat{s}^2.
\end{eqnarray*}
Therefore, the empirical and self-average mathematical relations are respectively 
\begin{eqnarray*}
M(\phi,\hat{m},\hat{s})
&=& \frac{1}{2}\lambda^2 \hat{s}\hat{m}^2+(-\lambda \hat{s}\hat{m}^2/2+\mu-\varepsilon \hat{s}/2)^2
\\
&&
+(-\lambda  \hat{s}\hat{m}^2/2+\mu/2-\varepsilon \hat{s}/2)(1+ \hat{s}^2M_4)
- \lambda +\lambda \hat{m}  \hat{s}^2M_3, \\
\langle M \rangle (\phi,\hat{m},\hat{s})
&=&
 \frac{1}{2}\lambda^2 \hat{s}\hat{m}^2+(-\lambda \hat{s}\hat{m}^2/2+\mu-\varepsilon \hat{s}/2)^2
\\
&&
+4(-\lambda  \hat{s}\hat{m}^2/2+\mu/2-\varepsilon \hat{s}/2) - \lambda .
\end{eqnarray*} 
When $\lambda=\varepsilon=0$, $M(\varphi,\hat{m},\hat{s})$ is minimized at $\mu=-(1+\hat{s}^2M_4)/4$,
whereas $\langle M\rangle(\phi,\hat{m},\hat{s})$ at $\mu=-1$.

In this model, we can derive the exact forms of CV, WAIC, DIC, and the free energy, hence we can compare
the optimal hyperparameters for these criteria. Let 
\[
Z_n(X,\alpha)=\int p(X|w)^{\alpha}\prod_{i=1}^n p(X_i|w)\varphi(w)dw.
\]
Then
\[
Z_n(X,\alpha)= \frac{1}{(2\pi)^{(n+\alpha-1)/2}}\exp\Bigl(-\frac{\log a(\alpha)}{2}-c(\alpha)\log d(\alpha)\Bigr)\Gamma(c(\alpha)),
\]
where $\Gamma(\;\;)$ is the gamma function and 
\begin{eqnarray*}
a(\alpha)&=&\alpha+\lambda+n, \\
b_i(\alpha)&=&\alpha X+\sum_{j=1}^{n}X_j^2,\\
c(\alpha)&=&\mu+(\alpha+n+1)/2,\\
d_i(\alpha)&=&(1/2)(\alpha X+\sum_{j=1}^n X_j^2-b_i(\alpha)^2/a(\alpha)+\varepsilon).
\end{eqnarray*}
All criteria can be calculated by using $Z(X,\alpha)$ by their definitions, 
\begin{eqnarray*}
{\rm CV}(\varphi)&=&-\frac{1}{n}\sum_{i=1}^n\{\log Z_n(0,0)-\log Z_n(X_i,-1)\}, \\
{\rm WAIC}(\varphi)&=&-\frac{1}{n}\sum_{i=1}^n\{\log Z_n(X_i,1)-\log Z_n(0,0)-\frac{\partial^2}{\partial \alpha^2}(\log Z_n(X_i,0))\},\\
{\rm DIC}(\varphi)&=&-\frac{1}{n}\sum_{i=1}^n\{2 \frac{\partial}{\partial \alpha} (\log Z_n(X_i,0))-\log p(X_i,\overline{m},\overline{s})\},\\
F_{free}(\varphi)&=&-\log Z_n(0,0)+\log Z_0(0,0),
\end{eqnarray*}
where $\overline{m}=b(0)/a(0)$ and  $\overline{s}=(2\mu+n+1)/(\sum_iX_i^2-b(0)^2/a(0)+\varepsilon)$.

A numerical experiment was conducted.  A true distribution $q(x)$ was set as ${\cal N}(1,1^2)$. 
We study a case $n=25$. Ten thousands independent training sets were collected. 
A statistical model and a prior were 
 defined by eq.(\ref{eq:ND}) and eq.(\ref{eq:Prior}) respectively. The fixed prior was $\varphi_0(w)\equiv 1$. 
 We set $\lambda=\varepsilon=0.01$, and studied the optimization problem of the hyperparameter 
 $\mu$. Firstly, we compared averages and standard deviations of criteria. 
In Table.\ref{table:111}, averages and standard deviations  of 
 \begin{eqnarray*}
 \Delta C&=& {\rm CV}(\varphi)-{\rm CV}(\varphi_0), \\
  \Delta W&=&{\rm WAIC}(\varphi)-{\rm WAIC}(\varphi_0),\\
 {\rm WAICR}&=&M(\phi,\hat{w})/n^2,\\
{\rm WAICRS}&=&\langle M\rangle (\phi,\hat{w})/n^2,\\
  \Delta D&=&{\rm DIC}(\varphi)-{\rm DIC}(\varphi_0),\\
  \Delta G&=&G(\varphi)-G(\varphi_0),
 \end{eqnarray*}
are shown 
for the two cases $\mu=\pm 1$. In this experiment, averages of $\Delta C$, $\Delta W$, ${\rm WAICR}$, and ${\rm WAICRS}$ were
almost equal to that of $\Delta G$, however that of $\Delta D$ was not. 
The standard deviations were
\[
\sigma(\Delta G)>>\sigma(\Delta C)>\sigma(\Delta W)>\sigma({\rm WAICR})>\sigma({\rm WAICRS})\cong \sigma(\Delta DIC).
\]
The standard deviation of $\Delta G$ was largest which is consistent to Theorem 1. Note that CV had the larger variance 
than WAIC. WAICRS gave the most precise result. 
\begin{table}[tb]
\begin{center}
\begin{tabular}{|c|c|c|c|c|c|c|c|}
\hline
 &  $\mu$  &     $\Delta C$  &     $\Delta W$  &   WAICR   &  WAICRS  &  $\Delta D$   &    $\Delta G$ \\
\hline
Average & -1 & -0.00194 & -0.00175 & -0.00147 & -0.00165 & 0.00332 & -0.00156 \\
\hline
STD  & -1 & 0.00101 & 0.00080 & 0.00062 &  0.00001 & 0.00001 & 0.01292 \\
\hline
Average & 1 &0.00506 & 0.00489  & 0.00450   &0.00467  & 0.00006  & 0.00445\\
\hline
STD  & 1  &0.00095  & 0.00076  & 0.00059   &0.00004  & 0.00002   &0.01250\\
\hline
\end{tabular}
\end{center}
\caption{Averages and Standard Errors of Criteria}
\label{table:111}
\end{table}

Secondly, we compared the distributions of the chosen hyperparameters by criteria.  
One hundred candidate hyper parameters in the interval $(-2.5,2.5]$ were compared and 
the optimal hyperparameter for each criterion was chosen by minimization. 
Remark that the interval for the free energy was set as $(-0.5,2.5]$ because the prior is proper
if and only if $\mu>-0.5$. In Table.\ref{table:222},
averages (A), standard deviations (STD), and $A\pm 2STD$ of optimal hyperparameters are shown. 

In this case, the optimal hyperparameter for the minum  generalization loss is almost equal to $(-1)$, whose prior is
improper. 
By CV, WAIC, WAICR, WAICRS, the optimal hyperparameter was almost chosen, whereas by DIC or the free energy,
it was not. The standard deviations of chosen hyperparameters were
\begin{eqnarray*}
\sigma(h({\rm CV}))& > & \sigma(h({\rm WAIC}))>\sigma(h({\rm WAICR}))\\
&>& \sigma(h(F))>\sigma(h({\rm DIC}))>\sigma(h({\rm WAICRS})). 
\end{eqnarray*}
In this experiment, neither the marginal likelihood nor DIC was appropriate for the predictive prior design. 

\begin{table}[tb]
\begin{center}
\begin{tabular}{|c|c|c|c|c|c|c|}
\hline
   & h(CV )&       h(WAIC)    &  h(WAICR)  &   h(WAICRS) &   h(DIC)   &   h(F)\\
  \hline  
Average  &  -0.9863 &   -0.9416  &  -0.9329 &   -0.9993 &   0.4512 &   -0.2977\\
\hline
STD   &  0.2297  &  0.19231  &  0.1885  &  0.0059  &  0.0077  &  0.0106\\
\hline
A$-$2STD  &   -1.4456 &   -1.3262  &  -1.3100  &  -1.0112  &  0.4358  &  -0.3188\\
\hline
A$+$2STD  &  -0.5269  &  -0.5569  &  -0.5559  &  -0.9874 &   0.4667 &   -0.2766 \\
\hline
\end{tabular}
\end{center}
\caption{Chosen Hyperparameters in Normal distribution}
\label{table:222}
\end{table}

\subsection{Linear Regression} 

Let us study a linear regression problem. Let a statistical model of 
$y\in\RR^1$, $x\in \RR^d$, $w\in \RR^d$, and $\lambda\in\RR^1$ be 
\begin{eqnarray}
p(y|x,w)&=&\frac{1}{(2\pi\sigma^2)^{1/2}}\exp\Bigl(-\frac{1}{2\sigma^2}(y-w\cdot x)^2\Bigr),\label{eq:linear_p} \\
\varphi(w|\lambda)&=&\exp(-\frac{\lambda}{2}\|w\|^2),\label{eq:linear_varphi}
\end{eqnarray}
where 
$\sigma>0$ is a constant. The basic prior is set as $\varphi_0(w)\equiv 1$. Hence the log 
prior ratio function is $\phi(w)=\varphi(w|\lambda)$ and 
$\hat{w}$ is equal to MLE. The function $L(w)$ without a constant term is 
\begin{eqnarray}
L(w) &=&\frac{1}{2\sigma^2}\sum_{i=1}^{n}(y_i-w\cdot x_i)^2.
\end{eqnarray}
It is immediately derived that 
\begin{eqnarray}
\hat{w}_{k_1}&=&(\sum_{i=1}^{n} x_{ik_1}x_{ik_2})^{-1}(\sum_{i=1}^{n} y_i x_{ik_1}),\\
L_{k_1k_2}(\hat{w}) &=&\frac{1}{\sigma^2 n}\sum_{i=1}^nx_{ik_1}x_{ik_2}, \\
L_{k_1k_2k_3}(\hat{w})&=&0,\\
F_{k_1,k_2}(\hat{w})&=&\frac{1}{\sigma^4 n}\sum_{i=1}^n (y_i-\hat{w}\cdot x_i)^2x_{ik_1}x_{ik_2}, \\
F_{k_1k_2,k_3}(\hat{w})&=&-\frac{1}{\sigma^4 n}\sum_{i=1}^n (y_i-\hat{w}\cdot x_i)x_{ik_1}x_{ik_2}x_{ik_3}, \\
(\log\phi)_{k_1}(\hat{w})&=&-\lambda \hat{w}_{k_1},\\
(\log\phi)_{k_1k_2}(\hat{w})&=&-\lambda \delta_{k_1k_2},
\end{eqnarray}
and $J^{k_1k_2}(\hat{w})=(L_{k_1k_2})^{-1}(\hat{w})$. 
Hence
\begin{eqnarray}
A^{k_1k_2}(\hat{w})&=& \frac{1}{2}J^{k_1k_2}(\hat{w}),\\
B^{k_1k_2}(\hat{w})&=&\frac{1}{2}\{J^{k_1k_2}(\hat{w})+J^{k_1k_3}(\hat{w})J^{k_2k_4}(\hat{w})F_{k_3,k_4}(\hat{w})\},\\
C^{k_1}(\hat{w})&=&J^{k_1k_2}(\hat{w})J^{k_3k_4}(\hat{w})F_{k_2k_3,k_4}(\hat{w}),
\end{eqnarray}
resulting that 
\begin{eqnarray}
M(\phi,\hat{w})&=&\frac{\lambda^2}{2}(J^{k_1k_2}(\hat{w})\hat{w}_{k_1}\hat{w}_{k_2})
-\lambda \mbox{tr}(B(\hat{w})+ C^{k_1}(\hat{w})\hat{w}_{k_1}),\\
\langle M\rangle (\phi,\hat{w})&=&\frac{\lambda^2}{2}(J^{k_1k_2}(\hat{w})\hat{w}_{k_1}\hat{w}_{k_2})
-\lambda \mbox{tr}(J(\hat{w})),
\end{eqnarray}
where we used $\langle B\rangle^{k_1k_2}=J^{k_1k_2}$  and $\langle C\rangle^{k_1}(\hat{w})=0$. 
In this model the optimal hyperparameter for WAICRS is directly given by
\begin{equation}\label{eq:opt_lambda}
\lambda= \frac{\mbox{tr}(J(\hat{w}))}{J^{k_1k_2}(\hat{w})\hat{w}_{k_1}\hat{w}_{k_2}}.
\end{equation}
The exact CV, WAIC, DIC, and the free energy are also calculated. Let 
\[
Z_n(X,Y,\alpha)=\int p(Y|X,w)^{\alpha}\prod_{i=1}^n p(Y_i|X_i,w)\varphi(w)dw.
\]
Then
\[
Z_n(X,Y,\alpha)= \frac{(\lambda\sigma^2)^{d/2}
\exp\Bigl(\frac{1}{2\sigma^2}\{b(\alpha)^TA(\alpha)^{-1}b(\alpha)-(\alpha Y^2+\sum_{i=1}^nY_i^2)\}\Bigl)
}{(2\pi\sigma^2)^{(n+\alpha)/2}\det(A(\alpha))^{1/2}},
\]
where
\begin{eqnarray*}
A(\alpha)&=&\alpha XX^T+\sigma^2\lambda I + \sum_{i=1}^{n}X_iX_i^T,\\
b(\alpha)&=&\alpha YX+\sum_{i=1}^n Y_iX_i.
\end{eqnarray*}
All criteria can be calculated by using $Z(X,Y,\alpha)$ by their definitions, 
\begin{eqnarray*}
{\rm CV}(\varphi)&=&-\frac{1}{n}\sum_{i=1}^n\{\log Z_n(0,0,0)-\log Z_n(X_i,Y_i,-1)\}, \\
{\rm WAIC}(\varphi)&=&-\frac{1}{n}\sum_{i=1}^n\{\log Z_n(X_i,Y_i,1)-\log Z_n(0,0,0)-\frac{\partial^2}{\partial \alpha^2}(\log Z_n(X_i,Y_i,0))\},\\
{\rm DIC}(\varphi)&=&-\frac{1}{n}\sum_{i=1}^n\{2 \frac{\partial}{\partial \alpha} (\log Z_n(X_i,0,0))-\log p(X_i,Y_i,\overline{m})\},\\
F_{free}(\varphi)&=&-\log Z_n(0,0,0)+\log Z_0(0,0,0),
\end{eqnarray*}
where $\overline{m}=A(0)^{-1}b(0)$.

A numerical experiment was conducted for the case $q(x,y)=q(x)p(y|x,w_0)$. Here
$q(x)$ was  the 
normal distribution ${\cal N}(a_0,I_5)$, where $a_{0}=(1,1,1,1,1)$, and $I_5$ is the $d=5$ dimensional
identity matrix, and $w_0=(1,1,1,1,1)$. 
A constant $\sigma=0.1$ was set. The candidate hyperparameters for $\lambda $ were
taken from the interval $(0,10)$. Distributions of chosen hyperparameters are shown in Table.\ref{table:333}. 
The average hyperparmeters chosen by ${\rm CV}$, ${\rm WAIC}$, ${\rm WAICR}$, ${\rm WAICRS}$, and ${\rm DIC}$ were almost equal to each others.
The variances by WAICRS and DIC were smaller than other methods. The optimal hyperparameter by the marginal likelihood 
was different from other methods. In this case, the posterior distribution is rigorously equal to the normal 
distribution and the true distribution is realizable by a statistical model, hence DIC can be applied, whose value was almost 
equal to WAICRS. Note that, in this model, the true parameter is $w_0=0$, then the optimal hyperparameter $\lambda$ 
diverges as $n\rightarrow\infty$. This phenomenon is caused by the fact that $w_0=0$ is contained in the divergent parameter,
which is discussed in Section \ref{section:div}.
\begin{table}[tb]
\begin{center}
\begin{tabular}{|c|c|c|c|c|c|c|}
\hline
   & h(CV )&       h(WAIC)    &  h(WAICR)  &   h(WAICRS) &   h(DIC)   &   h(F)\\
  \hline  
Average  & 5.0064  &  5.0017  &  4.9320 &   5.0253  &  5.0248 &  1.0000\\
\hline
STD   &  1.9358  & 1.9297  & 1.8808  & 0.2960  & 0.2961  & 0.0000 \\
\hline
\end{tabular}
\end{center}
\caption{Chosen Hyperparameters in linear regression}
\label{table:333}
\end{table}

\section{Basic Lemmas}

The main purpose of this paper is to prove Theorems 1 and 2. 
In this section we prepare several lemmas which are used in the proof of the main theorem. 

For arbitrary function $f(w)$,  we define the expectation values by 
\begin{eqnarray*}
\EE_{\varphi}^{(\pm j)}[f(w)]&=&
\frac{\displaystyle
\int f(w)\varphi(w) p(X_j|w)^{\pm 1}\prod_{i=1}^n p(X_i|w)dw
}{\displaystyle
\int \varphi(w)p(X_j|w)^{\pm 1}\prod_{i=1}^n p(X_i|w)dw
}.
\end{eqnarray*}
Then the predictive distribution of $x$ using training samples $X^n$ leaving $X_j$ out is 
\[
\EE_{\varphi}^{(-j)}[p(x|w)].
\]
Thus its log loss for the test sample $X_j$ is
\[
-\log \EE_{\varphi}^{(-j)}[p(X_j|w)].
\]
The log loss of the leave-one-out cross validation is then given by 
\begin{equation}\label{eq:cvcv}
{\rm CV}(\varphi)=-\frac{1}{n}\sum_{j=1}^n \log \EE_{\varphi}^{(-j)}[p(X_j|w)].
\end{equation}

\begin{Lemma}
Let $\phi(w)=\varphi(w)/\varphi_0(w)$. The cross validation and the generalization error
satisfy the following equations. 
\begin{equation}
{\rm CV}(\varphi)=
{\rm CV}(\varphi_0)+
\frac{1}{n}\sum_{j=1}^n\{
\log \EE_{\varphi_0}^{(-j)}[\phi(w)]
-\log \EE_{\varphi_0}[\phi(w)]
\},
\end{equation}
\begin{equation}
G(\varphi)=
G(\varphi_0)-
\EE_{X_{n+1}}\Bigl[
\log \EE_{\varphi_0}^{(+(n+1))}[\phi(w)]
-\log \EE_{\varphi_0}[\phi(w)]\Bigr].
\end{equation}
\end{Lemma}
(Proof of Lemma 1)
By the definitions $\EE_{\varphi}[\;\;]$, $\EE_{\varphi}^{(- j)}[\;\;]$,  it follows that 
\begin{eqnarray*}
\EE_{\varphi}^{(-j)}[p(X_j|w)]&=&
\frac{\displaystyle
\int \varphi(w) \prod_{i=1}^n p(X_i|w)dw
}{\displaystyle
\int \varphi(w)\prod_{i\neq j}^n p(X_i|w)dw
} \\
&=& 
\frac{\displaystyle
\int \varphi(w)\prod_{i=1}^n p(X_i|w)dw
}{\displaystyle
\int \varphi_0(w)\prod_{i=1}^n p(X_i|w)dw
} 
\frac{\displaystyle
\int \varphi_0(w)\prod_{i=1 }^n p(X_i|w)dw
}{\displaystyle
\int \varphi_0(w)\prod_{i\neq j}^n p(X_i|w)dw
} \\
&&\times\frac{\displaystyle
\int \varphi_0(w)\prod_{i\neq j }^n p(X_i|w)dw
}{\displaystyle
\int \varphi(w)\prod_{i\neq j}^n p(X_i|w)dw
} 
\\
&=&\EE_{\varphi_0}[\phi]\;\;\EE_{\varphi_0}^{(-j)}[p(X_j|w)]\;/\;\EE_{\varphi_0}^{(-j)}[\phi].
\end{eqnarray*}
By the definition ${\rm CV}(\varphi)$, eq.(\ref{eq:cvcv}), the first half of Lemma 1 is obtained. 
For the latter half, 
\begin{eqnarray*}
\EE_{\varphi}[p(X_{n+1}|w)]&=&
\frac{\displaystyle
\int \varphi(w)p(X_{n+1}|w)\prod_{i=1}^n p(X_i|w)dw
}{\displaystyle
\int \varphi(w)\prod_{i=1}^n p(X_i|w)dw
} \\
&=& 
\frac{\displaystyle
\int \varphi(w)\prod_{i=1 }^{n+1} p(X_i|w)dw
}{\displaystyle
\int \varphi_0(w)\prod_{i=1}^{n+1} p(X_i|w)dw
} 
\frac{\displaystyle
\int \varphi_0(w)\prod_{i=1 }^{n+1} p(X_i|w)dw
}{\displaystyle
\int \varphi_0(w)\prod_{i=1}^n p(X_i|w)dw
} \\
&&\times\frac{\displaystyle
\int \varphi_0(w)\prod_{i=1 }^n p(X_i|w)dw
}{\displaystyle
\int \varphi(w)\prod_{i=1}^n p(X_i|w)dw
} 
\\
&=&\EE_{\varphi_0}^{+(n+1)}[\phi]\;\;\EE_{\varphi_0}[p(X_{n+1}|w)]\;/\;\EE_{\varphi_0}[\phi].
\end{eqnarray*}
By using the definition of the generalization error, eq.(\ref{eq:G(varphi)}), 
the latter half of Lemma 1 is obtained. (Q.E.D.) 
\vskip3mm
\noindent{\bf Definition}.
The log loss function for  $X^n\setminus X_j$  is defined by 
\[
L(w,-j)=-\frac{1}{n}\sum_{i\neq j}^n\log p(X_i|w)-\frac{1}{n}\log \varphi_0(w).
\]
The MAP estimator for $X^n\setminus X_j$ is denoted by 
\[
\check{w}_j=\mbox{arg}\min L(w,-j).
\]

\begin{Lemma} 
Let $f(w)$ be a function $Q(X^n,w)$ which satisfies the regularity conditions (1), (2), ..., (5). 
Then there exist  functions $R_1(f,w)$ and $R_2(f,w)$ which satisfy 
\begin{eqnarray}
\EE_{\varphi_0}[f(w)]&=& f(\hat{w})+\frac{R_1(f,\hat{w})}{n}+\frac{R_2(f,\hat{w})}{n^2}+O_p(\frac{1}{n^3}),\\
\EE_{\varphi_0}^{(-j)}[f(w)]&=& f(\check{w}_j)+\frac{R_1(f,\check{w}_j)}{n-1}+\frac{R_2(f,\check{w}_j)}{(n-1)^2}
+O_p(\frac{1}{n^3}),
\end{eqnarray}
where  $R_1(f,w)$ is given by
\begin{eqnarray}
R_1(f,w)&=&
 \frac{1}{2}f_{k_1k_2}(w)J^{k_1k_2}(w)-\frac{1}{2}f_{k_1}(w)V^{k_1} (w). 
 \end{eqnarray}
and $V^{k_1}(w)=J^{k_1k_2}(w)J^{k_3k_4}(w)L_{k_2k_3k_4}(w)$. 
\end{Lemma}
We do not need 
the concrete form of $R_2(f,w)$ in the proof of the main theorem. However,
it is given in the proof of Lemma 2, eq.(\ref{eq:R_2}). 
\vskip3mm\noindent
(Proof of Lemma 2) 
Since $L(\hat{w})$ is a constant function of $w$, by using the regularity condition (5), 
\begin{eqnarray}
 \EE_{\varphi_0}[f(w)]&=&
\frac{\int f(w)\exp(-nL(w)+nL(\hat{w}))dw}
{\int \exp(-nL(w)+nL(\hat{w}))dw} \nonumber \\
&=& f(\hat{w})+\frac{\int (f(w)-f(\hat{w}))\exp(-nL(w)+nL(\hat{w}))dw}
{\int \exp(-nL(w)+nL(\hat{w}))dw}\nonumber  \\
&=& f(\hat{w})+\frac{Z_1}{Z_0}(1 +O_p(\frac{1}{n^{\beta}})),
\end{eqnarray}
where 
\begin{eqnarray}
Z_1&=&\int_{W(\varepsilon)} (f(w)-f(\hat{w}))\exp(-nL(w)+nL(\hat{w}))dw,\label{eq:Z1Z1}\\
Z_0&=&\int_{W(\varepsilon)} \exp(-nL(w)+nL(\hat{w}))dw.\label{eq:Z0Z0}
\end{eqnarray}
Note that the definition of $W(\varepsilon)$ is given in eq.(\ref{eq:Wepsilon}). 
Let $u=\sqrt{n}(w-\hat{w})$. Then $du=n^{d/2}dw$ and the integrated region is $|u|<n^{\epsilon}$. 
The Taylor expansions of $f(w)$ and $nL(w)$ among $\hat{w}$ are 
respectively given by
\begin{eqnarray*}
f(w)-f(\hat{w})&=&H_1(u),\\
n(L(w)-L(\hat{w}))&=& \frac{1}{2}L_{k_1k_2}u^{k_1}u^{k_2}+ H_2(u) ,
\end{eqnarray*}
where $H_1(u)$ and $H_2(u)$ are functions defined by 
\begin{eqnarray}
H_1(u)&=& \frac{1}{n^{1/2}}f_{k_1}u^{k_1}+\frac{1}{2n}f_{k_1k_2}u^{k_1}u^{k_2}
+\frac{1}{6n^{3/2}}f_{k_1k_2k_3}u^{k_1}u^{k_2}u^{k_3} \nonumber \\
 &&+\frac{1}{24n^{2}}f_{k_1k_2k_3k_4}u^{k_1}u^{k_2}u^{k_3}u^{k_4} +\frac{g_1(u)}{n^{5/2}},\\
H_2(u)&=&\frac{1}{6n^{1/2}}L_{k_1k_2k_3}u^{k_1}u^{k_2}u^{k_3}
+\frac{1}{24n}L_{k_1k_2k_3k_4}u^{k_1}u^{k_2}u^{k_3}u^{k_4}\nonumber \\
&&+\frac{1}{120 n^{3/2}}L_{k_1k_2k_3k_4k_5}u^{k_1}u^{k_2}u^{k_3}u^{k_4}u^{k_5}+
\frac{g_2(u)}{n^2},\label{eq:H_2(u)}
\end{eqnarray}
where $g_1(u)$ and $g_{2}(u)$ are constant order functions. 
In these equations, the derivatives of $f$ and $L$ are defined by their values at $w=\hat{w}$. 
We use notations, 
\begin{eqnarray}
\rho(u)&=&\exp(-\frac{1}{2}L_{k_1k_2}u^{k_1}u^{k_2}),\\
c_0&=&\int \rho(u)du=\frac{(2\pi)^{d/2}}{\det(L_{k_1k_2})^{1/2}}.
\end{eqnarray}
Remark that $J^{k_1k_2}$ is the inverse matrix of $L_{k_1k_2}$, hence 
\begin{eqnarray}
\int u^{k_1}u^{k_2}\rho(u)du &=& c_0J^{k_1k_2}, \label{eq:u2}\\
\int  \prod_{j=1}^4u^{k_j}
\rho(u)du 
&=&c_0 \{J^{k_1k_2}J^{k_3k_4}+J^{k_1k_3}J^{k_2k_4}+J^{k_1k_4}J^{k_2k_3}\},\label{eq:u4} \\
 \int \prod_{j=1}^6u^{k_j}
\rho(u)du 
&=&c_0  Sym(J)^{k_1k_2k_3k_4k_5k_6},\label{eq:u6}\\
 \int \prod_{j=1}^8u^{k_j}
\rho(u)du 
&=&c_0  Sym(J)^{k_1k_2k_3k_4k_5k_6k_7k_8},\label{eq:u8}
\end{eqnarray}
where 
\begin{eqnarray}
Sym(J)^{k_1k_2k_3k_4k_5k_6}&=& \sum_{m_1,...,m_6} J^{m_1m_2}J^{m_3m_4}J^{m_5m_6},\\
Sym(J)^{k_1k_2k_3k_4k_5k_6k_7k_8}&=&\sum_{m_1,...,m_8} J^{m_1m_2}J^{m_3m_4}J^{m_5m_6}J^{m_7m_8}.
\end{eqnarray}
Here 
 $\sum_{m_1,...,m_6}$ is the sum of all $15$ different pair combinations of $(k_1,...,k_6)$
and $\sum_{m_1,...,m_8}$ is the sum of all $105$ different pair combinations of $(k_1,...,k_8)$.
By using these results, $Z_0$ in eq.(\ref{eq:Z0Z0}) is given by 
\begin{eqnarray}
Z_0
&=&\frac{1}{n^{d/2}}\int_{|u|<n^{\epsilon}}\exp(-H_2(u))\rho(u)du\nonumber \\
&=&\frac{1}{n^{d/2}} \int_{|u|<n^{\epsilon}} \{1-H_2(u)+\frac{H_2(u)^2}{2}-
\frac{H_2(u)^3}{6}+O_p(n^{-2})\}\rho(u)du.
\end{eqnarray}
Then by using the symmetry of the integrated region, the integrations of the odd order terms are equal to zero.
It follows that 
\begin{eqnarray}\label{eq:Z_0}
Z_0&=& \frac{c_0}{n^{d/2}}\Bigl(1+\frac{Y_1(\hat{w})}{n}+O_p(\frac{1}{n^2})\Bigr),
\end{eqnarray}
where
\begin{eqnarray}
Y_1(\hat{w})&=&
\frac{1}{c_0}
\int \{-\frac{1}{24}L_{k_1k_2k_3k_4}u^{k_1}u^{k_2}u^{k_3}u^{k_4} 
+\frac{1}{72}(L_{k_1k_2k_3}u^{k_1}u^{k_2}u^{k_3})^2
\}\rho(u)du\nonumber \\
&=&- \frac{1}{8}L_{k_1k_2k_3k_4}J^{k_1k_2}J^{k_3k_4} 
+\frac{5}{24}L_{k_1k_2k_3}L_{k_4k_5k_6}J^{k_1k_2}J^{k_3k_4}J^{k_5k_6},
\end{eqnarray}
where we used eq.(\ref{eq:u4}) and eq.(\ref{eq:u6}). 
On the other hand, $Z_1$ in eq.(\ref{eq:Z1Z1}) is given by 
\begin{eqnarray}
Z_1
&=&\frac{1}{n^{d/2}}\int_{|u|<n^{\epsilon} } H_1(u)\exp(-H_2(u))\rho(u)du\nonumber \\
&=&\frac{1}{n^{d/2}} \int_{|u|<n^{\epsilon} }   H_1(u)\{1-H_2(w)+\frac{H_2(w)^2}{2}-
\frac{H_2(u)^3}{6}+O_p(n^{-2})\}\rho(u)du.
\end{eqnarray}
Then by using symmetry of the integrated region, it follows that
\begin{eqnarray}
Z_1&=&  \frac{c_0}{n^{d/2}}\Bigl(\frac{Y_2(\hat{w})}{n}+
\frac{Y_3(\hat{w})}{n^2}+
O_p(\frac{1}{n^2})\Bigr),
\end{eqnarray}
where $Y_2(\hat{w})$ and $Y_3(\hat{w})$ are given by 
\begin{eqnarray}
Y_2(\hat{w})&=&
 \frac{1}{c_0}
\int \{\frac{1}{2}f_{k_1k_2}u^{k_1}u^{k_2}-\frac{1}{6}f_{k_1}L_{k_2k_3k_4}
u^{k_1}u^{k_2}u^{k_3}u^{k_4}\}\rho(u)du\nonumber  \\
&=& \frac{1}{2}f_{k_1k_2}J^{k_1k_2}-\frac{1}{2}f_{k_1}L_{k_2k_3k_4}
J^{k_1k_2}J^{k_3k_4},\\
Y_3(\hat{w})&=& \frac{1}{c_0}
\int \rho(u)du\Bigl\{\frac{1}{24}f_{k_1k_2k_3k_4}u^{k_1}u^{k_2}u^{k_3}u^{k_4} \nonumber\\
&&-(\frac{1}{120}f_{k_1}L_{k_2k_3k_4k_5k_6}
+\frac{1}{48}f_{k_1k_2}L_{k_3k_4k_5k_6}+\frac{1}{36}f_{k_1k_2k_3}L_{k_4k_5k_6})
\prod_{j=1}^6 u^{k_j} \nonumber\\
&&+\frac{1}{144}(f_{k_1k_2}L_{k_3k_4k_5}L_{k_6k_7k_8}+f_{k_1}L_{k_2k_3k_4}L_{k_5k_6k_7k_8})
\prod_{j=1}^8 u^{k_j}\Bigr\} \nonumber\\
&=&
\frac{1}{8}f_{k_1k_2k_3k_4}J^{k_1k_2}J^{k_3k_4}
-\Bigl(\frac{1}{120}f_{k_1}L_{k_2k_3k_4k_5}\nonumber \\
&&+\frac{1}{48}f_{k_1k_2}L_{k_1k_2k_3k_4}+\frac{1}{36}f_{k_1k_2k_3}L_{k_4k_5k_6}\Bigr)
Sym(J)^{k_1k_2k_3k_4k_5k_6}\nonumber \\
&&+\frac{1}{144}(f_{k_1k_2}L_{k_3k_4k_5}L_{k_6k_7k_8}+f_{k_1}L_{k_2k_3k_4}L_{k_5k_6k_7k_8})
Sym(J)^{k_1k_2k_3k_4k_5k_6k_7k_8},\nonumber 
\end{eqnarray}
where we used eq.(\ref{eq:u4}), eq.(\ref{eq:u6}), and eq.(\ref{eq:u8}). 
Summing up these results, 
\begin{eqnarray}
\EE_{\varphi_0}[f(w)]&=&  f(\hat{w})+\frac{Z_1}{Z_0}\Bigl(1 +O_p(\frac{1}{n^{\beta}})\Bigr) \nonumber\\
&=&  f(\hat{w})+\frac{Y_2(\hat{w})/n+Y_3(\hat{w})/n^2+
O_p(1/n^{3})}{1+Y_1(\hat{w})/n+O_p(1/n^{2})} (1 +O_p(\frac{1}{n^{\beta}})) \nonumber\\
&=&  f(\hat{w})+Y_2(\hat{w})/n+(Y_3(\hat{w})-Y_1(\hat{w})Y_2(\hat{w}))/n^2+
O_p(1/n^{3}).
\end{eqnarray}
Therefore, by putting 
\begin{eqnarray}
R_1(f,\hat{w})&=&Y_2(\hat{w}), \\
R_2(f,\hat{w})&=& Y_3(\hat{w})-Y_1(\hat{w})Y_2(\hat{w}), \label{eq:R_2}
\end{eqnarray}
the first half of Lemma is completed. 
The latter half is equal to the case when the training samples are $X^n$ leaving $X_j$ out, hence
it is immediately obtained from the first half. (Q.E.D.)

\begin{Lemma} 
Let $|\alpha|\leq 1$. If $m$ is a positive odd number, 
\begin{equation}\label{eq:oddnumber}
\frac{\EE_{\varphi_0}[p(X_k|w)^{\alpha}\prod_{j=1}^{m}(w^{k_j}-\hat{w}^{k_j})]}
{\EE_{\varphi_0}[p(X_k|w)^{\alpha}]}
=O_p(\frac{1}{n^{(m+1)/2}}).
\end{equation}
If $m$ is a positive even number, 
\begin{equation}\label{eq:evennumber}
\frac{\EE_{\varphi_0}[p(X_k|w)^{\alpha}\prod_{j=1}^{m}(w^{k_j}-\hat{w}^{k_j})]}
{\EE_{\varphi_0}[p(X_k|w)^{\alpha}]}=O_p(\frac{1}{n^{m/2}}).
\end{equation}
For $m=2,4$, 
\begin{eqnarray}
&& \EE_{\varphi_0}[\prod_{j=1}^{2}(w^{k_j}-\hat{w}^{k_j})]=\frac{1}{n}J^{k_1k_2}+O_p(\frac{1}{n^2}), \label{eq:Lemma3m2}\\
&& \EE_{\varphi_0}[\prod_{j=1}^{4}(w^{k_j}-\hat{w}^{k_j})]=\frac{1}{n^2}(J^{k_1k_2}J^{k_3k_4}+J^{k_1k_3}J^{k_2k_4}
+J^{k_1k_4}J^{k_2k_3})
+O_p(\frac{1}{n^3}).  \label{eq:Lemma3m4}
\end{eqnarray}
\end{Lemma}
(Proof of Lemma 3)
In this proof, we use same notations as the proof of Lemma2. By the regularity condition (5), 
\[
\frac{\EE_{\varphi_0}[p(X_k|w)^{\alpha}\prod_{j=1}^{m}(w^{k_j}-\hat{w}^{k_j})]}
{\EE_{\varphi_0}[p(X_k|w)^{\alpha}]}
=\frac{Z_m^{*}}{Z_0^{*}}(1 +O_p(\frac{1}{n^{\beta}}))
\]
for an arbitrary $\beta>0$, where 
\begin{eqnarray}
Z_m^{*}&\equiv&  \int_{W(\epsilon)} \prod_{j=1}^m(w^{k_j}-\hat{w}^{k_j}) \exp(-nL(w)+nL(\hat{w})+\alpha \eta(X_k,w)) dw\nonumber \\
&=&\frac{1}{n^{(m+d)/2}} \int_{|u|<n^{\epsilon}}\prod_{j=1}^m u^{k_j} \exp(-H_2(u)+\alpha\eta(X_k,\hat{w}+u/\sqrt{n}))\rho(u)du.
\end{eqnarray}
Here we used a notation,  $\eta(X_k,w)=\log p(X_k|w)-\log p(X_k|\hat{w})$. 
By eq.(\ref{eq:H_2(u)}), the first term of $H_2(u)$ is in proportion to $u^3/n^{1/2}$.
Moreover, there exists $u^{*}$ such that 
$\eta(X_k,\hat{w}+u/\sqrt{n})=(u/\sqrt{n})\eta(X_k,u^{*})$.
 Hence by the expansion of 
 \[
 \exp(-H_2(u)+\alpha u \eta(X_k,u^{*})/\sqrt{n})=1+ \{ -H_2(u)+\alpha u \eta(X_k,u^{*})/\sqrt{n}\}+O_p(1/n), 
 \]
 if $m$ is an odd number, $Z_m=O_p(1/n^{(m+d+1)/2})$, which shows eq.(\ref{eq:oddnumber}), and 
 if $m$ is an even number, $Z_m=O_p(1/n^{(m+d)/2})$, which shows eq.(\ref{eq:evennumber}). 
By using eq.(\ref{eq:u2}) and eq.(\ref{eq:u4}) in the case $\alpha=0$, the results for $m=2,4$  are derived.  (Q.E.D.) 
\vskip3mm
\noindent{\bf Definition}. We use several functions of $w$  in the proof. 
\begin{eqnarray}
S^{k_1}(w)&=& J^{k_1k_2}(w)J^{k_3k_4}(w)F_{k_2k_3,k_4}(w),  \\
T^{k_1}(w)&=&J^{k_1k_2}(w)J^{k_3k_4}(w)J^{k_5k_6}(w)L_{k_2k_3k_5}(w)F_{k_4,k_6}(w), \\
U^{k_1k_2}(w)&=&J^{k_1k_3}(w)J^{k_2k_4}(w)F_{k_3,j_4}(w) .
\end{eqnarray}

\begin{Lemma}  
Let $\hat{w}$ and $\check{w}_j$ and be the MAP estimator for $X^n$ and $X^n\setminus X_j$,
respetivelly. Then 
\begin{eqnarray}
&& \frac{1}{n}\sum_{j=1}^n\{(\check{w}_j)^{k_1}-\hat{w}^{k_1}\}=
\frac{1}{n^2}S_{k_1}(\hat{w})-\frac{1}{2n^2}T_{k_1}(\hat{w})+O_p(\frac{1}{n^3}),
\label{eq:Lemma4_1} \\
&& \frac{1}{n}\sum_{j=1}^n\{(\check{w}_j)^{k_1}-\hat{w}^{k_1}\}\{(\check{w}_j)^{k_2}-\hat{w}^{k_2}\}=
\frac{1}{n^2}U^{k_1k_2}(\hat{w})+O_p(\frac{1}{n^3}).\label{eq:Lemma4_2}
\end{eqnarray}
For an arbitrary $C^{\infty}$-class function $f(w)$
\begin{eqnarray}
\frac{1}{n}\sum_{j=1}^{n}\{f(\check{w}_j)-f(\hat{w})\}&=& 
\frac{1}{n^2}\{f_{k_1}(\hat{w})(S^{k_1}(\hat{w})-\frac{1}{2}T^{k_1}(\hat{w}))\nonumber \\
&& +\frac{1}{2}f_{k_1k_2}(\hat{w})U^{k_1k_2}(\hat{w})\}+O_p(\frac{1}{n^3}). 
\label{eq:Lemma4_3}
\end{eqnarray}
\end{Lemma}
\noindent
(Proof of Lemma 4)
In this proof, we use a notation, $\ell(j,w)=\log p(X_j,w)$. 
Since $\check{w}_j$ minimizes $L(w,-j)$, its derivative is equal to zero at $\check{w}_j$, 
\begin{eqnarray}\label{eq:L30}
L_{k_1}(\check{w}_j,-j)=0.
\end{eqnarray}
By using the mean value theorem, there exists a parameter $w^{*}$ which satisfies
$\|w^{*}-\hat{w}\|\leq \|\check{w}_j-\hat{w}\|$ and 
\begin{eqnarray}\label{eq:L31}
L_{k_1}(\hat{w},-j)+L_{k_1k_2}(w^{*},-j)((\check{w}_j)^{k_2}-\hat{w}^{k_2})=0. 
\end{eqnarray}
Note that the MAP estimator $\hat{w}$ minimizes $L(w)$ where 
\begin{eqnarray}\label{eq:L31.5}
L(w)=L(w,-j)-\frac{1}{n}\ell(j,w),
\end{eqnarray}
hence its derivative satisfies $L_{k_1}(\hat{w})=0$. Therefore 
\begin{eqnarray}\label{eq:L32}
L_{k_1}(\hat{w},-j)=\frac{1}{n}\ell_{k_1}(j,\hat{w}). 
\end{eqnarray}
By applying eq.(\ref{eq:L32}) to eq.(\ref{eq:L31}), 
\begin{eqnarray}
(\check{w}_j)^{k_1}-\hat{w}^{k_1}=-\frac{1}{n}(L(w^{*},-j)^{-1})^{k_1k_2}\ell_{k_1}(j,\hat{w}). 
\end{eqnarray}
Therefore $\check{w}_j-\hat{w}=O_p(1/n)$, resulting that  $w^{*}-\hat{w}=O_p(1/n)$. 
It follows that 
\begin{eqnarray}
(\check{w}_j)^{k_1}-\hat{w}^{k_1}&=& -\frac{1}{n}(L(\hat{w},-j)^{-1})^{k_1k_2}\ell_{k_2}(j,\hat{w})+O_p(1/n^2)\nonumber \\ 
&=& -\frac{1}{n}(L(\hat{w})^{-1})^{k_1k_2}\ell_{k_2}(j,\hat{w})+O_p(1/n^2),\label{eq:L32.5}
\end{eqnarray}
where we used $L(w,-j)=L(w)+O_p(1/n)$. 
Hence
\begin{eqnarray}
&& \frac{1}{n}\sum_{j=1}^{n}
((\check{w}_j)^{k_1}-\hat{w}^{k_1})
((\check{w}_j)^{k_2}-\hat{w}^{k_2})\nonumber \\
&&=
\frac{1}{n^3}\sum_{j=1}^{n}(L(\hat{w})^{-1})^{k_1k_3}\ell_{k_3}(j,\hat{w})
(L(\hat{w})^{-1})^{k_2k_4}\ell_{k_4}(j,\hat{w})+O_p(1/n^3) \nonumber \\
&&=
\frac{1}{n^2}J^{k_1k_3}J^{k_2k_4}F_{k_3,k_4}+O_p(1/n^3) ,
\end{eqnarray}
which shows eq.(\ref{eq:Lemma4_2}) in Lemma 4.@Let us show  eq.(\ref{eq:Lemma4_1}).
From eq.(\ref{eq:L30}), 
by using the higher order mean value theorem, there exists a parameter $w^{**}$ which satisfies
$\|w^{**}-\hat{w}\|\leq \|\check{w}_j-\hat{w}\|$ and 
\begin{eqnarray}
&& \frac{1}{n}\ell_{k_1}(j,\hat{w})+L_{k_1k_2}(\hat{w},-j)((\check{w}_j)^{k_2}-\hat{w}^{k_2}) \nonumber \\
&& +\frac{1}{2}L_{k_1k_2k_3}(w^{**},-j)((\check{w}_j)^{k_2}-\hat{w}^{k_2})((\check{w}_j)^{k_3}-\hat{w}^{k_3})=0,
\label{eq:L33}
\end{eqnarray}
where we used eq.(\ref{eq:L32}). The second term of eq.(\ref{eq:L33}) is 
\begin{eqnarray}
&& L_{k_1k_2}(\hat{w},-j)((\check{w}_j)^{k_2}-\hat{w}^{k_2}) \nonumber \\
&& = L_{k_1k_2}(\hat{w})((\check{w}_j)^{k_2}-\hat{w}^{k_2}) 
+  \frac{1}{n}\ell_{k_1k_2}(j,\hat{w})((\check{w}_j)^{k_2}-\hat{w}^{k_2})\nonumber \\
&&= L_{k_1k_2}(\hat{w})((\check{w}_j)^{k_2}-\hat{w}^{k_2})  
-\frac{1}{n^2}\ell_{k_1k_2}(j,\hat{w})(L(\hat{w})^{-1})^{k_2k_3}\ell_{k_3}(j,\hat{w})+O_p(1/n^3),\label{eq:L34}
\end{eqnarray}
where we used eq.(\ref{eq:L31.5}) and eq.(\ref{eq:L32.5}). Also by eq.(\ref{eq:L32.5}), the third term of eq.(\ref{eq:L33}) is 
\begin{eqnarray}
&& \frac{1}{2}L_{k_1k_2k_3}(w^{**},-j)((\check{w}_j)^{k_2}-\hat{w}^{k_2})((\check{w}_j)^{k_3}-\hat{w}^{k_3}) \nonumber\\
&&=  \frac{1}{2}L_{k_1k_2k_3}(\hat{w})((\check{w}_j)^{k_2}-\hat{w}^{k_2})((\check{w}_j)^{k_3}-\hat{w}^{k_3})
+O_p(\frac{1}{n^3})\nonumber \\
&& = \frac{1}{2n^2}L_{k_1k_2k_3}(\hat{w})
(L(\hat{w})^{-1})^{k_2k_4}\ell_{k_4}(j,\hat{w})
(L(\hat{w})^{-1})^{k_3k_5}\ell_{k_5}(j,\hat{w})+O_p(\frac{1}{n^3}). \label{eq:L35}
\end{eqnarray}
Then by applying eq.(\ref{eq:L34}), eq.(\ref{eq:L35}),  $L_{k_1}(\hat{w})=0$, 
and $(L^{-1})^{k_1k_2}=J^{k_1k_2}$ to eq.(\ref{eq:L33}), 
the sum for $j=1,2,...,n$ of eq.(\ref{eq:L33}) results in 
\begin{eqnarray}
&& L_{k_1k_2}\Bigl(\frac{1}{n}\sum_{j=1}^n ((\check{w}_j)^{k_2}-\hat{w}^{k_2}) \Bigr)
-\frac{1}{n^2}F_{k_1k_2,k_3}J^{k_2k_3}\nonumber\\
&& +\frac{1}{2n^2}L_{k_1k_2k_3}J^{k_2k_4}J^{k_3k_5}F_{k_4,k_5}+O_p(\frac{1}{n^3})=0.
\end{eqnarray}
Therefore 
\begin{eqnarray}
\frac{1}{n}\sum_{j=1}^n ((\check{w}_j)^{k_1}-\hat{w}^{k_1}) 
&=& \frac{1}{n^2}J^{k_1k_2}J^{k_3k_4}F_{k_2k_3,k_4}\nonumber\\
&& -\frac{1}{2n^2}J^{k_1k_2}J^{k_3k_4}J^{k_5k_6}L_{k_2k_3k_5}F_{k_4,k_6}+O_p(\frac{1}{n^3}),
\end{eqnarray}
which shows  eq.(\ref{eq:Lemma4_1}) in Lemma 4. The last equation in Lemma 4, eq.(\ref{eq:Lemma4_3}),  is proved by 
 \begin{eqnarray}
 \frac{1}{n}\sum_{j=1}^{n}\{f(\check{w}_j)-f(\hat{w})\}&=& 
  \frac{1}{n}\sum_{j=1}^{n}((\check{w}_j)^{k_1}-\hat{w}^{k_1})f_{k_1}\nonumber  \\
  && + \frac{1}{2n}\sum_{j=1}^{n}((\check{w}_j)^{k_1}-\hat{w}^{k_1})((\check{w}_j)^{k_2}-\hat{w}^{k_2})f_{k_1k_2} 
  +O_p(\frac{1}{n^3})\nonumber \\
  &=& \frac{1}{n^2}f_{k_1}(S^{k_1}-T^{k_1}/2)+\frac{1}{2n^2}f_{k_1k_2}U^{k_1k_2}  +O_p(\frac{1}{n^3}),
 \end{eqnarray}
which completes Lemma 4. (Q.E.D.)

\section{Proof of Theorem 1}

In this section, we prove the main theorems. The proof of Theorem 1 consists of the five parts, 
Cross validation, WAIC, mathematical relations, averages, and random generalization loss. 

\subsection{Proof of Theorem1, Cross Validation}

In this subsection, we prove eq.(\ref{eq:CV})  in Theorem 1. 
By Lemma 2,
\begin{eqnarray}
\EE_{\varphi_0}[\phi(w)]&=& \phi(\hat{w})
\Bigl(1+\frac{R_1(\phi,\hat{w})}{\phi(\hat{w})n}+\frac{R_2(\phi,\hat{w})}{\phi(\hat{w})n^2}\Bigr)+O_p(\frac{1}{n^3}),\\
\EE_{\varphi_0}^{(-j)}[\phi(w)]&=&
 \phi(\check{w}_j)\Bigl(1+\frac{R_1( \phi,\check{w}_j)}{\phi(\check{w}_j)(n-1)}+\frac{R_2( \phi,\check{w}_j)}{\phi(\check{w}_j)(n-1)^2}
 \Bigr)
+O_p(\frac{1}{n^3}),
\end{eqnarray}
For an arbitrary $C^{\infty}$-class function $f(w)$, by Lemma 4, 
\begin{eqnarray}
\frac{f(\check{w}_j)}{n-1}-\frac{f(\hat{w})}{n}&=& 
\frac{f(\check{w}_j)-f(\hat{w})}{n-1}+\frac{f(\hat{w})}{n(n-1)}\nonumber  \\
&=& \frac{f(\hat{w})}{n^2}+O_p(\frac{1}{n^3}),
\end{eqnarray}
\begin{eqnarray}
\frac{f(\check{w}_j)}{(n-1)^2}-\frac{f(\hat{w})}{n^2}&=& 
\frac{f(\check{w}_j)-f(\hat{w})}{(n-1)^2}+\frac{(2n-1)f(\hat{w})}{n^2(n-1)^2} \nonumber \\
&=& O_p(\frac{1}{n^3}).
\end{eqnarray}
By using Lemma 1 and by applying these equations for $f(w)=R_1(w)/\phi(w), R_2(w)/\phi(w)$,
\begin{eqnarray}
{\rm CV}(\varphi)-{\rm CV}(\varphi_0)&=&\frac{1}{n}\sum_{j=1}^n\{
\log \EE_{\varphi_0}^{(-j)}[\phi(w)]
-\log \EE_{\varphi_0}[\phi(w)] \}\nonumber \\
&=& \frac{1}{n}\sum_{j=1}^n\{\log\phi(\check{w}_j)-\log\phi(\hat{w})\} 
+ \frac{R_{1}(\hat{w})}{\phi(\hat{w})n^2}+O_p(\frac{1}{n^3}).
\end{eqnarray}
By using Lemma 2 and 4, 
\begin{eqnarray}
{\rm CV}(\varphi)-{\rm CV}(\varphi_0)
&=& \frac{1}{n^2}(\log\phi)_{k_1}(S^{k_1}-\frac{1}{2}T^{k_1})
+\frac{1}{2n^2}(\log\phi)_{k_1k_2}U^{k_1k_2} \nonumber \\
&& + \frac{1}{2\phi n^2}(\phi_{k_1k_2}J^{k_1k_2}-\phi_{k_1}V^{k_1})
+O_p(\frac{1}{n^3}).
\end{eqnarray}
Then by using
\begin{eqnarray*}
\phi_{k_1}/\phi&=& (\log\phi)_{k_1},\\
\phi_{k_1k_2}/\phi&=& (\log\phi)_{k_1k_2}+(\log\phi)_{k_1}(\log\phi)_{k_2},
\end{eqnarray*}
it follows that 
\begin{eqnarray}
{\rm CV}(\varphi)-{\rm CV}(\varphi_0)
&=& \frac{1}{n^2}(\log\phi)_{k_1}(S^{k_1}-\frac{1}{2}T^{k_1}-\frac{1}{2}V^{k_1}) \nonumber \\
&&+\frac{1}{2n^2}(\log\phi)_{k_1k_2}(U^{k_1k_2}+J^{k_1k_2}) \nonumber \\
&& + \frac{1}{2 n^2}(\log\phi)_{k_1}(\log\phi)_{k_2}J^{k_1k_2}
+O_p(\frac{1}{n^3}),
\end{eqnarray}
which completes eq.(\ref{eq:CV}).  (Q.E.D.)

\subsection{Proof of Theorem 1, WAIC}

In this subsection we prove eq.(\ref{eq:WAIC}) and eq.(\ref{eq:CV_WAIC}) in Theorem 1. 
In the following, we prove eq.(\ref{eq:CV_WAIC}).
In order to prove eq.(\ref{eq:CV_WAIC}), it is sufficient to prove eq.(\ref{eq:CV_WAIC}) in the case $\varphi(w)0=\varphi_0(w)$
for an arbitrary $\varphi_0(w)$. 
Let the functional cumulant generating function for $\varphi_0(w)$ be 
\[
F_{cum}^{0}(\alpha)=\frac{1}{n}\sum_{i=1}^{n}\log \EE_{\varphi_0}[p(X_i|w)^{\alpha}].
\]
For a natural number $j$, we define  the $j$th functional cumulant by 
\[
C_{j}(\alpha)\equiv \frac{\partial^{j}}{\partial \alpha^j}F_{cum}^0(\alpha).
\]
Then by definition, $F_{cum}^0(0)=0$ and 
\begin{eqnarray}
{\rm CV}(\varphi_0)&=&F_{cum}^0(-1),\label{eq:fcumcv}\\
{\rm WAIC}(\varphi_0)&=&T(\varphi_0)+V(\varphi_0)/n=-F_{cum}^0(1)+C_2(0).\label{eq:fcumwaic}
\end{eqnarray}
For a natural number $j$, let $m_j(X_i,\alpha)$ be 
\[
m_{j}(X_i,\alpha)=\frac{\EE_{\varphi_0}[\eta(X_i,w)^j\exp(-\alpha\eta(X_i,w))]}
{\EE_{\varphi}[\exp(-\alpha\eta(X_i,w))]},
\]
where $\eta(X_i,w)=\log p(X_i|w)-\log p(X_i|\hat{w})$. 
Note that 
\begin{equation}\label{eq:etaeta}
\eta(X_i,w)=(w^{k_1}-\hat{w}^{k_1})\ell_{k_1}(X_i,\hat{w})+O_p((w-\hat{w})^2).
\end{equation}
Therefore, if $j$ is an odd number, by using Lemma 3, 
\begin{equation}
m_j(X_i,\alpha)=O_p(\frac{1}{n^{(j+1)/2}}),\label{eq:mjodd}
\end{equation}
or if $j$ is an even number 
\begin{equation}
m_j(X_i,\alpha)=O_p(\frac{1}{n^{j/2}}).\label{eq:mjeven}
\end{equation}
Since $p(X_i|\hat{w})$ is a constant function of $w$, 
\begin{eqnarray}
C_6(\alpha)&=& \frac{1}{n}\sum_{i=1}^{n} \frac{\partial^{6}}{\partial \alpha^6}\log \EE_{\varphi_0}[p(X_i|w)^{\alpha}] \nonumber \\
&=& \frac{1}{n}\sum_{i=1}^{n} \frac{\partial^{6}}{\partial \alpha^6}\log \EE_{\varphi_0}[(p(X_i|w)/p(X_i|\hat{w}))^{\alpha}]\nonumber\\
&=& \frac{1}{n}\sum_{i=1}^{n} \frac{\partial^{6}}{\partial \alpha^6}\log \EE_{\varphi_0}[\exp(-\alpha\eta(X_i|w))]\nonumber\\
&=&\frac{1}{n}\sum_{i=1}^{n}\{m_6-6m_5m_1-15m_4m_2+30m_4m_1^2-10m_3^2+120m_3m_2m_1\nonumber \\
&&-120m_3m_1^3+30m_2^3-270m_2^2m_1^2+360m_2m_1^4-120m_1^6\}=O_p(\frac{1}{n^3}), \label{eq:mmm}
\end{eqnarray}
where $m_k=m_k(X_i,\alpha)$ in eq.(\ref{eq:mmm}). Hence by eq.(\ref{eq:fcumcv}),
\begin{eqnarray}
{\rm CV}(\varphi_0)&=&\sum_{j=1}^{5}\frac{(-1)^j}{j!}C_j(0)+O_p(\frac{1}{n^3}).
\end{eqnarray}
On the other hand, by eq.(\ref{eq:fcumwaic}),
\begin{eqnarray}
{\rm WAIC}(\varphi_0)=\sum_{j=1}^{5}\frac{-1}{j!}C_j(0)+C_2(0)+O_p(\frac{1}{n^3}).
\end{eqnarray}
It follows that 
\[
{\rm WAIC}(\varphi_0)={\rm CV}(\varphi_0)-\frac{1}{12}C_4(0)+O_p(\frac{1}{n^3}).
\]
Hence the main difference between ${\rm CV}$ and ${\rm WAIC}$ is $C_4(0)/12$.
In order to prove eq.(\ref{eq:CV_WAIC}), it is sufficient to prove $C_4(0)=O_p(1/n^3)$. 
\begin{eqnarray}
C_4(0)&=& \frac{1}{n}\sum_{i=1}^{n} \frac{\partial^{4}}{\partial \alpha^4}\log \EE_{\varphi}[p(X_i|w)^{\alpha}]\Bigl{|}_{\alpha=0} \nonumber \\
&=& \frac{1}{n}\sum_{i=1}^{n} \frac{\partial^{4}}{\partial \alpha^4}\log \EE_{\varphi}[(p(X_i|w)/p(X_i|\hat{w}))^{\alpha}]\Bigl{|}_{\alpha=0}\nonumber \\
&=&\frac{1}{n}\sum_{i=1}^n\{m_{4}-4m_3m_1-3m_{2}^2+12m_2m_{1}^2-6m_{1}^4\},\label{eq:mmmm}
\end{eqnarray}
where $m_k=m_k(X_i,0)$ in eq.(\ref{eq:mmmm}). By eq.(\ref{eq:mjodd}) and eq.(\ref{eq:mjeven}), 
\[
C_4(0)=\frac{1}{n}\sum_{i=1}^n\{
m_{4}(X_i,0)-3m_{2}(X_i,0)^2\}+O_p(\frac{1}{n^3}).
\]
By using a notation
\[
F_{k_1,k_2,k_3,k_4}\equiv \frac{1}{n}\sum_{i=1}^{n}\ell_{k_1}(X_i)\ell_{k_2}(X_i)\ell_{k_3}(X_i)\ell_{k_4}(X_i),
\]
it follows that by eq.(\ref{eq:etaeta}) and Lemma 3,
\begin{eqnarray}
\frac{1}{n}\sum_{i=1}^{n}m_4(X_i)&=&\frac{3}{n^2}J^{k_1k_2}J^{k_3k_4}
F_{k_1,k_2,k_3,k_4}
+O_p(1/n^3),\label{eq:uuuu}\\
\frac{1}{n}\sum_{i=1}^{n}m_2(X_i)^2&=&\frac{1}{n^2}J^{k_1k_2}J^{k_3k_4}
F_{k_1,k_2,k_3,k_4}+O_p(1/n^3),\label{eq:uu:uu}
\end{eqnarray}
resulting that $C_4(0)=O_p(1/n^3)$,  which completes eq.(\ref{eq:CV_WAIC}).  
 Then, eq.(\ref{eq:WAIC}) is immediately derived using eq.(\ref{eq:CV}) and eq.(\ref{eq:CV_WAIC}). (Q.E.D.) 

\subsection{Mathematical Relations between Priors}

In this subsection, we  prove eq.(\ref{eq:MM}), eq.(\ref{eq:MM2}), and eq.(\ref{eq:EMM}).  Since $\hat{w}$ minimizes $L(w)$, 
\[
L_{k_1}(\hat{w})=0.
\]
There exists $w^{*}$ such that $\|w^{*}-w_{0}\|\leq \|\hat{w}-w_0\|$ and that 
\[
L_{k_1}(w_0)+L_{k_1k_2}(w^{*})(\hat{w}-w_0)=0. 
\]
By the regularity condition (3), $\hat{w}\rightarrow w_0$ resulting that $w^{*}\rightarrow w_0$. 
By using the central limit theorem, 
\begin{equation}\label{eq:what}
\hat{w}-w_0=(L_{k_1k_2}(w^{*}))^{-1}L_{k_1}(w_0)=O_p(\frac{1}{\sqrt{n}}).
\end{equation}
Also by the central limit theorem, for an arbitrary $w\in W$, 
\begin{eqnarray}
L_{k_1k_2}(w)&=&\EE[L_{k_1k_2}(w)]+\frac{\beta_{k_1k_2}}{n^{1/2}},\label{eq:betak1k2}\\
L_{k_1k_2k_3}(w)&=& \EE[L_{k_1k_2k_3}(w)]+\frac{\beta_{k_1k_2k_3}}{n^{1/2}},\\
F_{k_1,k_2}(w)&=&\EE[F_{k_1,k_2}(w)]+\frac{\gamma_{k_1k_2}}{n^{1/2}},\\
F_{k_1k_2,k_3}(w)&=& \EE[F_{k_1k_2,k_3}(w)]+\frac{\gamma_{k_1k_2k_3}}{n^{1/2}},
\end{eqnarray}
where $\beta_{k_1k_2}$, $\beta_{k_1k_2k_3}$, $\gamma_{k_1k_2}$ 
and $\gamma_{k_1k_2k_3}$ are constant order random variables, whose expectation values are equal to zero.
By the definitions, 
\begin{eqnarray}
{\cal L}_{k_1k_2}(w)&=&\EE[L_{k_1k_2}(w)]+O(\frac{1}{n}),\label{eq:calLk1k2}\\
{\cal L}_{k_1k_2k_3}(w)&=& \EE[L_{k_1k_2k_3}(w)]+O(\frac{1}{n}),\\
{\cal F}_{k_1,k_2}(w)&=&\EE[F_{k_1,k_2}(w)]+O(\frac{1}{n}),\\
{\cal F}_{k_1k_2,k_3}(w)&=& \EE[F_{k_1k_2,k_3}(w)]+O(\frac{1}{n}),
\end{eqnarray}
Let $\beta\equiv \{\beta_{k_1k_2}\}$ and  $\Lambda\equiv \{L_{k_1k_2}(w)\}$. Then by eq.(\ref{eq:betak1k2}) and 
eq.(\ref{eq:calLk1k2}), 
\begin{eqnarray}
{\cal J}(w)&=& \EE[\Lambda]^{-1}+O(\frac{1}{n})\nonumber\\
&=& (\Lambda-\beta/\sqrt{n})^{-1}+O(\frac{1}{n})\nonumber\\
&=& (\Lambda(1-\Lambda ^{-1}\beta /\sqrt{n}))^{-1}+O(\frac{1}{n})\nonumber\\
&=& (1+\Lambda ^{-1}\beta /\sqrt{n})\Lambda^{-1}+O_p(\frac{1}{n})\nonumber\\
&=& J(w)+\Lambda^{-1}\beta\Lambda^{-1}/\sqrt{n}+O_p(\frac{1}{n}).\nonumber
\end{eqnarray}
Hence 
\begin{eqnarray*}
J^{k_1k_2}(w)&=&{\cal J}^{k_1k_2}(w)+O_p(1/\sqrt{n}),\\
\EE[J^{k_1k_2}(w)]&=&{\cal J}^{k_1k_2}(w)+O(1/n).
\end{eqnarray*}
It follows that 
\begin{eqnarray*}
M(\phi,w_0)&=& {\cal M}(\phi,w_0)+O_p(1/\sqrt{n}),\\
\EE[M(\phi,w_0)]&=& {\cal M}(\phi,w_0)+O(1/n).
\end{eqnarray*}
Hence
\begin{eqnarray*}
M(\phi,\hat{w})&=&M(\phi,w_0)+(\hat{w}-w_0)^{k_1}(M(\phi,w_0))_{k_1}={\cal M}(\phi,w_0)+O_p(\frac{1}{\sqrt{n}}),\\
\EE[M(\phi,\hat{w})]&=&{\cal M}(\phi,w_0)+O(\frac{1}{n}),
\end{eqnarray*}
which shows eq.(\ref{eq:MM}) and eq.(\ref{eq:EMM}). Then eq.(\ref{eq:MM2}) is immediately derived by the fact
$\hat{w}-\EE_{\varphi}[w]=O_p(1/n)$ by Lemma 3.
(Q.E.D.) 

\subsection{Proof of Theorem 1, Averages}

 In this subsection, we prove we show eq.(\ref{eq:ECV}), eq.(\ref{eq:EWAIC}), and eq.(\ref{eq:EG}).
 
Firstly, eq.(\ref{eq:ECV}) is derived from eq.(\ref{eq:CV}) and eq.(\ref{eq:EMM}). 
Secondly, eq.(\ref{eq:EWAIC}) is derived from eq.(\ref{eq:WAIC}) and eq.(\ref{eq:EMM}). 
Lastly, let us prove eq.(\ref{eq:EG}). 
Let ${\rm CV}_{n}(\varphi)$ and $G_n(\varphi)$ be the cross validation and 
the generalization losses for $X^n$, respectively. Then by the definition, for an arbitrary
$\varphi$, 
\begin{eqnarray}
\EE[G_n(\varphi)]&=& \EE[{\rm CV}_{n+1}(\varphi)]\nonumber \\
&=& \EE[{\rm CV}_{n+1}(\varphi_0)]+\frac{\EE[M(\varphi,\hat{w})]}{(n+1)^2}+O(\frac{1}{n^3}) \nonumber\\
&=& \EE[G_n(\varphi_0)]+\frac{\EE[M(\varphi,\hat{w})]}{n^2}+O(\frac{1}{n^3}) ,
\end{eqnarray}
where we used $1/n^2-1/(n+1)^2=O(1/n^3)$, which completes eq.(\ref{eq:EG}). (Q.E.D.) 

\subsection{Proof of Theorem1, Random Generalization Loss}

In this subsection, we prove eq.(\ref{eq:G}) in Theorem 1. 
We use a notation $\ell(n+1,w)=\log p(X_{n+1}|w)$. 
Let $\bar{w}$ be the parameter that minimizes
\[
-\frac{1}{n+1}\sum_{i=1}^{n+1}\log p(X_i|w)-\frac{1}{n+1}\log\varphi(w)=\frac{n}{n+1}\Bigl\{ L(w)-\frac{1}{n}\ell(n+1,w)\Bigr\}.
\]
Since $\bar{w}$ minimizes $L(w)-\ell(n+1,w)/n$,
\begin{equation}\label{eq:proveG1}
L_{k_1}(\bar{w})-\frac{1}{n}\ell_{k_1}(n+1,\bar{w})=0.
\end{equation}
By applying the mean value theorem to eq.(\ref{eq:proveG1}), there exists $\bar{w}^{*}$ such that 
\[
L_{k_1}(\hat{w})+(\bar{w}^{k_2}-\hat{w}^{k_2})
L_{k_1k_2}(\bar{w}^{*})-\frac{1}{n}\ell_{k_1}(n+1,\bar{w})=0.
\]
By using $L_{k_1}(\hat{w})=0$ and positive definiteness of $L_{k_1k_2}(\hat{w})$, 
\begin{equation}\label{eq:proveG2}
\bar{w}^{k_1}-\hat{w}^{k_1}=O_p(\frac{1}{n}).
\end{equation}
By applying the higher order mean value theorem to  eq.(\ref{eq:proveG1}), there exists $w^{**}$ such that 
\[
(\bar{w}^{k_2}-\hat{w}^{k_2})
L_{k_1k_2}(\hat{w})
+\frac{1}{2}(\bar{w}^{k_2}-\hat{w}^{k_2})(\bar{w}^{k_3}-\hat{w}^{k_3})
L_{k_1k_2k_3}(\hat{w}^{**})
-\frac{1}{n}\ell_{k_1}(n+1,\bar{w})=0.
\]
By eq.(\ref{eq:proveG2}), the second term of this equation is $O_p(1/n^2)$. The inverse matrix of $L_{k_1k_2}(\hat{w})$ is
$J^{k_1k_2}(\hat{w})$, 
\begin{equation}\label{eq:wk1hatwk1}
\bar{w}^{k_1}-\hat{w}^{k_1}=\frac{1}{n}J^{k_1k_2}(\hat{w})\ell_{k_2}(n+1,\hat{w})
+O_p(\frac{1}{n^2}).
\end{equation}
By eq.(\ref{eq:what}), $\hat{w}-w_0=O_p(1/\sqrt{n})$. Hence by the expansion of $(\hat{w}-w_0)$, 
\begin{eqnarray}
&& \EE_{X_{n+1}}[\ell_{k_2}(n+1,\hat{w})]=
\EE_{X_{n+1}}[(\log p(X_{n+1}|\hat{w}))_{k_2}]\nonumber\\
&&=(\hat{w}^{k_3}-(w_0)^{k_3})\EE_{X_{n+1}}[(\log p(X_{n+1}|w_0))_{k_2k_3}]+O_p(\frac{1}{n}),
\label{eq:ellk2}
\end{eqnarray}
where we used $\EE_{X_{n+1}}[(\log p(X_{n+1}|w_0))_{k_2}] =0$. By eq.(\ref{eq:wk1hatwk1}) and eq.(\ref{eq:ellk2}),
\begin{eqnarray}
\EE_{X_{n+1}}[\bar{w}^{k_1}-\hat{w}^{k_1}]
&=&-\frac{1}{n}J^{k_1k_2}(\hat{w})
(\hat{w}^{k_3}-(w_0)^{k_3})\EE[L_{k_2k_3}(w_0)]+O_p(\frac{1}{n^2})\nonumber \\
&=&-\frac{1}{n}(\hat{w}^{k_1}-(w_0)^{k_1})+O_p(\frac{1}{n^2}),\nonumber 
\end{eqnarray}
where we used $J^{k_2k_3}(\hat{w})=(\EE[L_{k_2k_3}(w_0)])^{-1}+O_p(1/n^{1/2})$.
Therefore, by Lemma 1 and 2
\begin{eqnarray}
G(\varphi)-G(\varphi_0)&=& -\EE_{X_{n+1}}[\log \EE_{\varphi_0}^{(+(n+1))}[\phi(w)]
-\log \EE_{\varphi_0}[\phi(w)]]\nonumber \\
&=& -\EE_{X_{n+1}}[-\log (\phi(\bar{w})+\frac{R_1(\phi,\bar{w})}{n+1})+\log (\phi(\hat{w})+\frac{R_1(\phi,\hat{w})}{n})]+O_p(\frac{1}{n^2})\nonumber  \\
&=& -\EE_{X_{n+1}}[(\bar{w}^{k_1}-\hat{w}^{k_1})](\log\phi)_{k_1}(\hat{w})+O_p(\frac{1}{n^2}) \nonumber  \\
&=& \frac{1}{n}
(\hat{w}^{k_1}-(w_0)^{k_1})(\log\phi)_{k_1}(\hat{w})+O_p(\frac{1}{n^2}) .
\end{eqnarray}
By eq.(\ref{eq:what}), $\hat{w}-w_0=O_p(1/\sqrt{n})$,  we obtain eq.(\ref{eq:G}). (Q.E.D.) 

\subsection{Proof of Theorem 2}

If there exists a parameter which satisfies $q(x)=p(x|w_0)$, then 
$\hat{w}-w_0=O_p(1/\sqrt{n})$, 
$\langle L_{k_1k_2}\rangle(w)={\cal L}_{k_1k_2}(w)+O_{p}(1/\sqrt{n})$,
$\langle L_{k_1k_2k_3}\rangle(w)={\cal L}_{k_1k_2}(w)+O_{p}(1/\sqrt{n})$,
$\langle F_{k_1,k_2}\rangle(w)={\cal F}_{k_1k_2}(w)+O_{p}(1/\sqrt{n})$, and 
$\langle F_{k_1k_2,k_3}\rangle(w)={\cal F}_{k_1k_2,k_3}(w)+O_{p}(1/\sqrt{n})$.
Hence Theorem 2 is obtained. (Q.E.D.)

\section{Discussions}

In this chapter, we discuss several points about predictive prior design. 

\subsection{Summary of Results}

In this paper, we have shown the mathematical properties of Bayesian CV, WAIC, and the generalization loss. 
Let us summarize the results of this paper. \\
(1) Even if the posterior distribution is not normal or even if the true distribution is unrealizable by a 
statistical model, CV and WAIC are applicable to predictive prior design. 
Theoretically CV and WAIC are asymptotically equivalent, whereas
experimentally the variance of WAIC is a little smaller than CV. 
In the regularity conditions are satisfied, then CV and WAIC can be approximated by 
WAICR. The variance of WAICR is a little smaller than CV and WAIC. \\
(2) If the true distribution is realizable by a statistical model, 
then CV and WAIC can be estimated by WAICRS. The variance of WAICRS is very smaller than WAICR. \\
(3) If the posterior distribution is rigorously normal and if the true distribution is realizable by a statistical
model, then DIC is almost equal to WAICRS. If otherwise, then DIC is different from CV, WAIC, or WAICRS
and the chosen hyperparameter by DIC is not optimal for predictive prior design in general. \\
(4) The marginal likelihood is not appropriate for predictive prior design.

\subsection{Divergence Phenomenon of CV and WAIC} \label{section:div}

In this subsection we study a divergence phenomenon of CV, WAIC, and the marginal likelihood. 
Let the maximum likelihood estimator be $w_{mle}$ and $\delta_{mle}(w)=\delta(w-w_{mle})$. 
Then for an arbitrary proper prior $\varphi(w)$, ${\rm CV}(\varphi)\geq {\rm CV}(\delta_{mle})$,
${\rm WAIC}(\varphi)\geq {\rm WAIC}(\delta_{mle})$, and $F_{free}(\varphi)\geq F_{free}(\delta_{mle})$. 
Hence, if a candidate prior can be made to converge to $\delta_{mle}(w)$, then minimizing these 
criteria results in the maximum likelihood method, where Theorem 1 does not hold. 

Assume that a proper prior $\varphi(w)=\varphi(w|\lambda)$ has a hyperparameter $\lambda$.
The set of divergent parameters of $\varphi(w|\lambda)$ is defined by 
\[
W_{div}(\varphi)=\{\overline{w}\in W\;;\;\mbox{ There exists } \{\lambda_k\} \mbox{ s.t. }
  \lim_{k\rightarrow\infty}\varphi(w|\lambda_k)\rightarrow \delta(w-\overline{w})\}.
\]
For example, if $\varphi(w|\lambda)=\sqrt{\lambda/2\pi}\exp(-\lambda w^2/2)$, then $W_{div}(\varphi)=\{0\}$, 
because $\lambda_k=k$ gives a sequence of priors which converges to the delta function.

If the optimal parameter $w_0$ that minimizes the average generalization loss is contained in $W_{div}(\varphi)$, then 
the optimal hyperparameter does not remain in a compact set 
as $n\rightarrow\infty$. For example, the optimal hyperparameter for WAICRS in eq.(\ref{eq:opt_lambda})
diverges if $\hat{w}\rightarrow 0$. In such cases, CV or WAIC may not have any minimum point, 
then the hyperparameter can not be optimized by using CV or WAIC. 
In such cases, some hyperprior or regularization term which is necessary. It is an important study to clarify 
the set of divergent parameters of a given prior. For example, if the Dirichlet distribution on $a \in (0,1)$
\[
Dir(a|\lambda_1,\lambda_2)= C(\lambda_1,\lambda_2) a^{\lambda_1}(1-a)^{\lambda_2},
\]
is used as a prior, then an arbitrary parameter in $(0,1)$ is contained in $W_{div}(Dir)$, because
\[
Dir(a| b_0 k, c_0 k)\rightarrow \delta(a-b_0/(b_0+c_0))\;\;\;(k\rightarrow\infty).
\]

\subsection{Training and Testing Sets}

In practical applications of machine learning, we often prepare both a set of training samples $X^n$ and
a set of test samples $Y^m$, where $X^n$ and $Y^m$ are independent. This method is sometimes
called the holdout cross validation. 
Then we have a basic question, 
``Does the optimal hyperparameter chosen by CV or WAIC using a training set $X^n$ 
minimize the generalization loss estimated using a test set $Y^n$ ?''. 
The theoretical answer to this question is No, 
because, as is shown in Theorem 1, the optimal hyperparameter 
for $X^n$ asymptotically minimizes $\EE[G(X^n)]$ but not $G(X^n)$. 
If one would find the hyperparameter which minimizes $G(X^n)$, then 
neither CV, WAIC, DIC, nor the marginal likelihood is appropriate. 
On the other hand, if one wants to measure the optimality of the chosen hyperparameter by
the criterion $\EE[G(X^n)]$, then
CV or WAIC using $Y^m$ is useful, because the optimal hyperparameter 
using $X^n$ is asymptitically equal to that using $Y^m$. 

\subsection{Self-Averaging and Bootstrap}

In this paper, we have shown  that the optimal hyperparameter for the minimum
average generalization loss can be found by CV and WAIC asymptotically. 
However, the variance of the estimated hyperparameter is sometimes not small. 
In experiments, WAICRS has very smaller variances than CV and WAIC. 
Although WAICRS can be used only in the case when the true distribution is realizable 
by a statistical model, it may be useful by its small variance. 

If a statistical model $p(x|w)$  is complicated and if it is difficult to derive the mathematical form of WAICRS,
then it can be estimated numerically by 
\[
{\rm WAICRS}\approx \EE_{Y^n}[{\rm WAIC}(Y^n,\varphi)-{\rm WAIC}(Y^n,\varphi_0)]
\]
where $Y^n$ is taken from the Bayesian predictive distribution $p^{*}(x)$ and ${\rm WAIC}(Y^n,\varphi)$ is WAIC for a set $Y^n$. 
Moreover, if $Y^n$ is taken from the empirical distribution $(1/n)\sum\delta(x-X_i)$, then 
the above equation approximates WAICR. These numerical methods may need heavy computational costs, however,
they may be useful if the precise hyperparamer optimization is necessary.

\section{Conclusion}

In this paper, we studied several methods how to design the hyperparameter 
from the predictive point of view. The mathematical relation between priors 
gives the explicit criterion of the prior and its variance is made smaller 
by using self-averaging. To construct the generalized theory of this paper onto singular
statistical model is the important problem for future study.

\subsection*{Acknowledgement}
This research was partially supported by the Ministry of Education,
Science, Sports and Culture in Japan, Grant-in-Aid for Scientific
Research 23500172 and 25120013.

\end{document}